\documentclass{article}

\usepackage[preprint,nonatbib]{neurips_2020}

\usepackage[utf8]{inputenc} 
\usepackage[T1]{fontenc}    
\usepackage{hyperref}       
\usepackage{url}            
\usepackage{booktabs}       
\usepackage{amsfonts}       
\usepackage{nicefrac}       
\usepackage{microtype}      
\usepackage{amssymb}
\usepackage{amsmath}
\usepackage{algorithm}
\usepackage{algorithmic}
\usepackage{graphicx,subcaption}
\usepackage{xcolor}

\newcommand{\pir}{\pi_r}

\newcommand{\pdk}[1]{p_d^{(#1)}}
\newcommand{\pidk}[1]{p_{id}^{(#1)}}
\newcommand{\xsupk}[2]{#1^{(#2)}}

\newcommand{\Qhstar}{\hat{Q}^*}
\newcommand{\Vhstar}{\hat{V}^*}

\title{Real-world Ride-hailing Vehicle Repositioning using Deep Reinforcement Learning \thanks{This paper is a significantly extended version of \cite{jtq2020repos}.}}

\author{%
  Yan Jiao \\
  DiDi Labs \\
  Mountain View, CA 94043 \\
  \texttt{yanjiao@didiglobal.com} \\
   \And
   Xiaocheng Tang \\
   DiDi Labs \\
   Mountain View, CA 94043 \\
   \texttt{xiaochengtang@didiglobal.com} \\
   \AND
   Zhiwei (Tony) Qin\thanks{corresponding author}\\
   DiDi Labs \\
   Mountain View, CA 94043 \\
   \texttt{qinzhiwei@didiglobal.com} \\
   \And
   Shuaiji Li \\
   DiDi Labs \\
   Mountain View, CA 94043 \\
   \texttt{shuaijili@didiglobal.com} \\
   \And
   Fan Zhang \\
   Didi Chuxing \\
   Beijing, China \\
   \texttt{feynmanzhangfan@didiglobal.com} \\
   \And
   Hongtu Zhu \\
   Didi Chuxing \\
   Beijing, China \\
   \texttt{zhuhongtu@didiglobal.com} \\
   \And
   Jieping Ye \\
   University of Michigan \\
   Ann Arbor, MI \\
   \texttt{jieping@gmail.com} \\
}

\begin{document}

\maketitle

\begin{abstract}
We present a new practical framework based on deep reinforcement learning and decision-time planning for real-world vehicle repositioning on ride-hailing (a type of mobility-on-demand, MoD) platforms.  
Our approach learns the spatiotemporal state-value function using a batch training algorithm with deep value networks.  The optimal repositioning action is generated on-demand through value-based policy search, which combines planning and bootstrapping with the value networks.  For the large-fleet problems, we develop several algorithmic features that we incorporate into our framework and that we demonstrate to induce coordination among the algorithmically-guided vehicles.
We benchmark our algorithm with baselines in a ride-hailing simulation environment to demonstrate its superiority in improving income efficiency measured by income-per-hour.  We have also designed and run a real-world experiment program with regular drivers on 
a major ride-hailing platform.  We have observed significantly positive results on key metrics comparing our method with experienced drivers who performed idle-time repositioning based on their own expertise.
\end{abstract}


\section{Introduction}
The emergence of ride-hailing platforms, led by companies such as DiDi, Uber, and Lyft, has revolutionized the way that personal mobility needs are met. As urban populations continue to grow in world's largest markets \cite{chinastats2019,michigan2019mod}, the current modes of transportation are increasingly insufficient to cope with the growing and changing demand. The digital platforms offer possibilities of much more efficient on-demand mobility by leveraging more global information and real-time supply-demand data. Auto industry experts expect that ride-hailing apps would eventually make individual car ownership optional, leading towards subscription-based services and shared ownership \cite{reuters2019ownership}. However, operational efficiency is a tough challenge for ride-hailing platforms, e.g., passenger waiting time can still be long \cite{wtop2019wait}, while drivers often chase surge based on their own experience which may not be always effective \cite{nyt2017surge}, and ride-hailing vehicles can be vacant 41\% of the time in a large city \cite{wsj2020vacant}. 

Vehicle repositioning and order matching are two major levers to improve the system efficiency of ride-hailing platforms. Vehicle repositioning \cite{shou2019find,lin2018efficient,oda2018movi} has direct influence on driver-side metrics and is an important lever to reduce driver idle time and increase the overall efficiency of a ride-hailing system, by proactively deploying idle vehicles to a specific location in anticipation of future demand at the destination or beyond. It is also an integral operational component of an autonomous ride-hailing system, where vehicles are fully managed by the dispatching algorithm.
On the other hand, order dispatching \cite{xu2018large,qin2018dispatching,tang2019deep,ozkan2017dynamic,korolko2018dynamic} matches idle drivers (vehicles) to open trip orders, which is a related but different problem.
Order dispatching directly impacts passenger-side metrics, e.g. order response rate and fulfillment rate. Together, these two levers collectively aim to align supply and demand better in both spatial and temporal spaces.

In this paper, we tackle the problem of vehicle repositioning on a ride-hailing platform, motivated by the development of an intelligent driver assistant program for helping the drivers on the platform to achieve better income. There are two scenarios for vehicle repositioning: small and large fleets. Both have their specific use cases. In the small-fleet problem, our objective is to learn an optimal policy that maximizes an individual driver's cumulative income rate, measured by \emph{income-per-hour} (IPH, see \eqref{eq:ind_iph}). This scenario can target, for example, those who are new to a ride-hailing platform to help them quickly ramp up by providing learning-based idle-time cruising strategies. This has significant positive impact to driver satisfaction and retention. Such program can also be used as bonus to incentivize high quality service that improves passenger ridership experience. Of course, our proposed repositioning algorithm is equally applicable to the management of a small autonomous vehicle fleet within a large-scale ride-hailing system. In the large-fleet case, on the other hand, we are interested in optimizing the IPH at the group level (see \eqref{eq:agg_iph}). This applies to the more general use case of managing the ride-hailing vehicles within an entire region (city). Care has to be taken to make sure that decisions on individual vehicles do not cause macro-level problems, for example, repositioning too many idle vehicles to a single high-demand spot.

We have implemented our proposed algorithm as the decision engine for an AI driver assistant program on a major ride-hailing platform and performed real-world field tests for both small and large fleets that demonstrate its superiority over human expert strategies.

\subsection{Related Works}
Many of the works on vehicle repositioning or taxi dispatching are under the setting of autonomous MoD, where a fleet of autonomous vehicles \cite{pendleton2017perception} are deployed in an MoD system and fully managed by a controller.  

\emph{Model predictive control} (MPC) and \emph{receding horizon control} (RHC) \cite{iglesias2018data,miller2017predictive,zhang2016model,miao2016taxi} are a popular class of methods that involve repeatedly solving a mathematical program using predictions of future costs and system dynamics over a moving time horizon to choose the next control action. 
A related method for fleet management in the trucking industry is \cite{simao2009approximate} where approximate dynamic programming (ADP) is used.
Previous works are all at grid-level granularity (with the solution only specifying the number of vehicles needed to transfer between cells) and using discrete time buckets with a short planning horizon (small number of buckets) due to computational complexity. For example, \cite{cheng18taxis} has a planning horizon of 30 minutes with 5-minute time steps. With the objective of finding the shortest route to the immediate next trip, a multi-arm bandits (MAB) method with Monte Carlo tree search \cite{garg2018route} has also been proposed. Our method differentiates in that it has a much longer optimization horizon (whole day) by using state-value functions and can easily meet the run-time requirement of our driver assistant application. 

A number of recent works use RL and deep RL to learn vehicle repositioning policies, such as dynamic programming \cite{shou2019find} and Monte Carlo learning \cite{verma2017augmenting}.  Through training within a grid-based simulation environment where vehicle within a grid are indistinguishable and matching happens only within a grid, DQN \cite{oda2018movi,holler2019deep}, multi-agent RL \cite{lin2018efficient}, and hierarchical RL \cite{jin2019coride} have been proposed for the repositioning problem, and in some cases, jointly with order dispatching.

For order matching/dispatching on ride-hailing platforms, a related but different problem from the vehicle repositioning problem, extensive literature has been proposed on, e.g., dynamic matching \cite{zhang2017taxi,korolko2018dynamic}, multi-stage stochastic optimization \cite{lowalekar2018online}, neural ADP for carpooling \cite{shah2020neural}, network flow optimization with rolling horizon \cite{bertsimas2019online}, and (deep) reinforcement learning \cite{xu2018large,qin2018dispatching,tang2019deep,li2019efficient}.

Vehicle repositioning and fleet management are also related to classical vehicle routing problems, where machine learning methods with deep neural networks have been used as new approaches to the TSP and the VRP, under an encoding-decoding framework \cite{nazari2018reinforcement,bello2016neural,vinyals2015pointer}.

\subsection{Algorithm Design Considerations}
In designing the solution framework for the vehicle repositioning problem, there are a few factors for consideration.  First of all, the algorithm has to be practical for implementation on real-world production system and for real-time execution.  That means we would not be able to make significant simplification assumptions that many some prior works have done.  In particular, grid-based algorithms with low decision granularity (i.e., only determining the cell-to-cell reposition quantities) and those require training or planning in a simulation environment (e.g., \cite{holler2019deep,garg2018route}) are hard to deploy in our setting. Second, our objective is to maximize an agent's daily income rate, which is long-term reward optimization on the scale of trips.  MPC with long horizon is expensive to solve.  A coarse discretization for the time would render the solution hard to implement and execute in our driver assistant program where review intervals are short (e.g., 100s).   On the other hand, RL, which focuses on long-term values, is well-suited for such objectives. We show later in simulation experiments the advantage in IPH of our method over an MAB-based method \cite{garg2018route} without considering temporal dependency of the decisions. Third, data from regular ride-hailing system is usually incomplete regarding idle-time repositioning due to online-offline behavior.  It is thus hard to learn a state-action value function for repositioning directly from data using such method as Monte Carlo learning \cite{verma2017augmenting}.

Considering the factors discussed above, we have developed a solution framework that combines offline batch RL and decision-time planning for guiding vehicle repositioning.
We model the problem within a semi-Markov decision process (semi-MDP) framework \cite{sutton1999between}, which optimizes a long-term cumulative reward (daily income rate) and models the impact of temporally extended actions (repositioning movements) on the long-term objective through state transitions along a policy.  
We learn the state value function using tailored spatiotemporal deep value networks  trained within a batch RL framework with dual policy evaluation.  We then use the state-value function and learned knowledge about the environment dynamics to develop a value-based policy search algorithm for real-time vehicle repositioning, which is a type of decision-time planning algorithm \cite{sutton2018reinforcement} and can be easily plugged into a \emph{generalized policy iteration} framework \cite{sutton2018reinforcement} for continuous improvement. For managing a large fleet, we simplify the search and focus more on the value function, taking into account the need for coordination among the vehicles.

\section{The Repositioning Problem}
We start from the environment dynamics.  The driver, when idle, can be assigned to an open trip order by the ride-hailing platform per the matching policy.  Order dispatching/matching takes place in a batch fashion typically with a time window of a few seconds \cite{tang2019deep,xu2018large,zhang2017taxi}.  
The driver goes to the origin of the trip (i.e. where the passenger is located) and transports the passenger to the destination.  
Trip fee is collected upon the completion of the trip.  
After dropping off the passenger, the driver becomes idle.  If the idle time exceeds a threshold of $L$ minutes (typically five to ten minutes), the driver performs repositioning by cruising to a specific destination, incurring a non-negative cost.  If the driver is to stay around the current location, he/she stays for $L$ minutes before another repositioning could be triggered.  During the course of any repositioning, the driver is still eligible for order assignment.  For modeling purpose, we assume that the driver can always complete a repositioning task before the platform would match the driver to an order, but this has little impact to the generality of the formulation, and the policy is not restricted by this assumption in any way. As we will explain below, a basic reposition moves the vehicle to a neighboring zone defined by a hexagonal grid system. As long as the zone structure is sufficiently granular, it is reasonable to assume that the reposition has been completed if the driver is dispatched in the destination zone. If the driver is dispatched while still in the origin zone, we simply treat it as if the reposition never takes place.

The objective of a repositioning algorithm is to maximize \emph{income efficiency} (or used interchangeably, \emph{income rate}), measured by income per (online) hour, IPH. This can be measured at an individual driver's level or at an aggregated level over a group of drivers. It is clear that vehicle repositioning is a sequential decision problem in which the current action affects the future income of the driver because of the spatiotemporal dependency.

\subsection{Semi-MDP Formulation}\label{subsec:formulation}
\begin{figure}
\begin{center}
    \hspace{-0in}
    \includegraphics[width=0.5\linewidth]{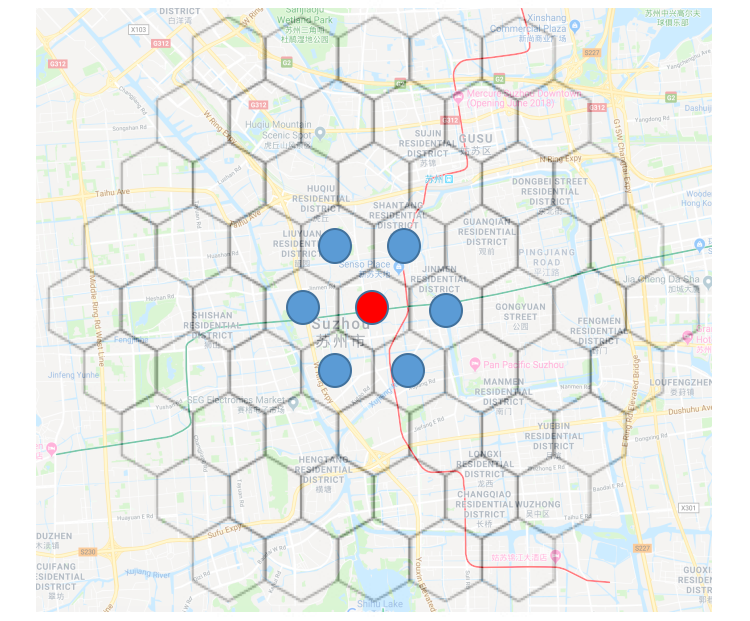}
	\caption{A sample map (obtained from Google Maps) overlaid with a hexagon grid system. The cell with a red circle represents the current cell of the vehicle, and those with blue circles are its neighboring cells.}
	\label{fig:hex_grids}
\end{center}
\end{figure}

Mathematically, we model the trajectory of a driver by a semi-MDP with the agent being the driver as follows.  
\paragraph{State} The driver's \emph{state} $s$ contains basic spatiotemporal information of location $l$ and time $t$, 
and it can include additional supply-demand contextual features $f$. Hence, we have $s=(l,t,f)$. 
\footnote{In Section \ref{sec:vps} and the associated experiments, only the basic spatiotemporal features $(l,t)$ are used. The contextual features $f$ are elaborated in Section \ref{sec:action-val}.} 
Although we use GPS coordinates for $l$, the basic geographical zones are defined by a hexagonal grid system, as illustrated in Figure
\ref{fig:hex_grids}. The hexagonal grid system is commonly used in
mapping systems because it has a desirable property that
the Euclidean distance between the center points of every
pair of neighboring grid cells is the same, and hexagonal
grids have the optimal perimeter/area ratio, which leads
to a good approximation of circles \cite{hales2001honeycomb}. 

\paragraph{Option}
The eligible actions for the agent include both vehicle repositioning and order fulfillment (as a result of order dispatching).  These actions are temporally extended, so they are \emph{options} in the context of a semi-MDP and are denoted by $o$.  A basic repositioning option is to go towards a destination in one of the six neighboring hexagonal cells of the driver plus the current cell itself.  See the cell with a red circle (current cell) and those with blue circles (neighboring cells) in Figure \ref{fig:hex_grids}.  
We use a single option $o_d$ to represent all the dispatching (order fulfillment) options, since in the context of this paper, the agent does not need to make dispatching decisions.
The time duration of a repositioning or order fulfillment option is $\tau_o$.  
\paragraph{Reward}
The price of the trip corresponding to an option is $p_o > 0$.  The cost of a repositioning option is $c_o \geq 0$.   Hence, the immediate reward of a transition is $r= -c_o$ for repositioning and $r=p_o$ for order fulfillment.  
The corresponding estimated version of $\tau_o$, $p_o$, and $c_o$ are $\hat{\tau}_o$, $\hat{p}_o$, and $\hat{c}_o$, respectively.  
\paragraph{Transition} The transition of the agent given a state and a repositioning option is deterministic, while the transition probability given a dispatching option $P(s'|s,o_d)$ is the probability of a trip going to $s'$ given it being assigned to the agent at $s$.

An \emph{episode} of this semi-MDP runs till the end of a day, that is, a state with its time component at midnight is terminal.
We denote the repositioning and the order dispatching policies separately by $\pir$ and $\pi_d$. At any decision point, either a dispatch or a reposition is performed to a vehicle. The vehicle is reviewed for order dispatching at highly granular time intervals (e.g., every two seconds). If it is matched, it starts to serve the order. Otherwise, it waits for the next review for dispatching or repositioning. Repositioning is reviewed at much coarser time intervals, e.g., every 100 seconds. If it is idle and not being reviewed for dispatching, $\pi_r$ determines a reposition for the vehicle. The overall policy $\pi$ can therefore be defined as 
\begin{equation}
    \pi(s) = 
    \begin{cases}
      \pi_r(s) & \text{if $s$ is a reposition review point,}\\
      \pi_d(s) & \text{otherwise (i.e., $s$ is a dispatch review point).}
    \end{cases}  
\end{equation}

In this paper, we are interested in only the reposition policy $\pi_r$ since dispatching is handled by an exogenous policy $\pi_{d0}$.  
\footnote{In fact, per our definition of the order dispatching option, $\pi_d$ is degenerate - there is only one option available.}
Hence, we would not explicitly learn $\pi_{d0}$, though we learn a state-value function associated with the current policy $\pi_0$ from the trip data as in Section \ref{sec:vnet}.

The state-option value function is denoted by $Q^{\pir}(s,o)$, with the understanding that it is also associated with $\pi_{d0}$.  $\hat{Q}$ denotes the approximation of the $Q$-function.  In the following sections, we develop a planning method to compute $\hat{Q}(s,o)$ for a particular $s$ so that the repositioning agent would be able to select the best movement at each decision point.  Our objective is to maximize the cumulative income rate (IPH), which is the ratio of the total price of the trips completed during an episode and the total online hours logged by a driver (individual level) or a group of drivers (group level). The individual-level IPH for a driver $x$ is defined as 
\begin{equation}\label{eq:ind_iph}
    p(x) := \frac{c(x)}{h(x)},
\end{equation}
where $c(\cdot)$ is the total income of the driver over the course of an episode, and $h(\cdot)$ is the total online hours of the driver. Similarly, the group-level IPH for a driver group $X$ is defined as
\begin{equation}\label{eq:agg_iph}
    p(X) := \frac{\sum_{x\in X}c(x)}{\sum_{x\in X}h(x)}.
\end{equation} 

\section{State Value-based Decision-time Planning}\label{sec:vps}
With the semi-MDP formulation in place, we first describe a repositioning strategy designed from the perspective of individual drivers who account for a small percentage of the driver population in a city and whose spatial activity ranges hardly interact. This approach falls in the family of decision-time planning, utilizing learned state values.

\subsection{Learning State Values}\label{sec:vnet}
\begin{figure}
\begin{center}
    \hspace{-0in}
    \includegraphics[width=0.7\linewidth]{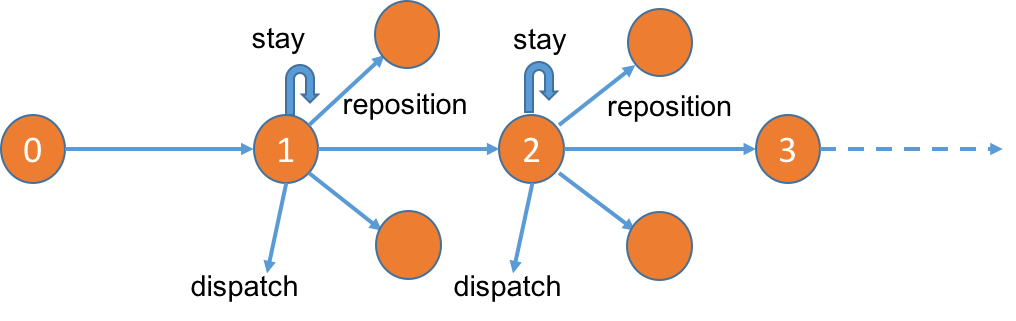}
	\caption{Environment dynamics model.  Any order dispatch happening during `reposition' or `stay' is assumed to take place at the next node. The nodes without marks represent the other eligible reposition destinations. The numbered nodes are repositions that the policy chooses.}
	\label{fig:env_model}
\end{center}
\end{figure}

We assume the following environment model per the dynamics and learn the state values associated with the semi-MDP above.  At state $s$, the probability of the driver being dispatched is $\pdk{s}$. The probability of being idle within the time interval of $L$ minutes is $\pidk{s} = 1 - \pdk{s}$, upon which repositioning is triggered next, and the driver can either stay around or move to another location.  If order dispatch takes place, the driver is relocated to the destination of the trip.  The estimated time interval for transitioning from the current state $s_0$ to the target state $s_i$ is $t_{0i} := \Delta t(s_0,s_i)$.  Transitions for both vehicle repositioning and order fulfillment are deterministic, given the current state and option.  This is illustrated in Figure \ref{fig:env_model}. 

We decompose the state value function into four components by conditioning on whether the driver is being dispatched or not, e.g.,
\begin{align}
\label{equ:v_decompose}
    V(s) = \pdk{s} V(s | dispatch) + \pidk{s} V(s | idle),
\end{align}
where $V(s | dispatch)$ and $V(s | idle)$ are the corresponding long-term value function conditioned on whether the associated driver is being dispatched or not at state $s$. In particular, we are interested in the conditional value function $V(s|dispatch)$. The reason will be clear in Section \ref{sec:planning}.

In our semi-MDP formulation, dispatching is represented by the single option $o_d$. So, $V(s|dispatch)$ is the conditional value function $V(s|o=o_d)$, and the dispatch probability $\pdk{s}$ is $\pi(o=o_d | s)$. To keep the notation concise, we will write $V(s|o=o_d)$ and $\pi(o=o_d | s)$ as $V(s|o_d)$ and $\pi(o_d|s)$ respectively.
Learning of \eqref{equ:v_decompose} can be done by evaluating the behavior policy on the observed transactions collected from the ride-hailing platform.
In particular, we apply variance reduction techniques to the joint learning of both $V(s)$ and the conditional value function $V(s |o_d)$. 
Here, we directly learn $V(s)$ from data instead of computing it using \eqref{equ:v_decompose} in order to minimize the errors due to estimation and composition.
The dispatch probabilities are estimated separately after applying negative sampling to drivers' trajectories.  
Next we describe the approach in details.

\subsubsection{Dual Policy Evaluation}\label{sec:dpe}
Notice that $V(s | o_d)$ is equivalent to the state-option value function $Q(s, o_d)$.
Evaluation of $V(s | o_d)$ can thus be done using standard TD algorithms like SARSA \cite{sutton2018reinforcement}.
In this work we propose Dual Policy Evaluation (DPE) that prevents stochasticity in the policy from increasing variance.
It does so by jointly learning $V(s | o_d)$ and $V(s)$ while basing the update of $V(s | o_d)$ not on the next-state value function $V(s' | o_d)$, but on the marginal expectation $V(s')$. 
Formally, consider the $k$-step transition from $s_0$ to $s_k$ by applying the option $o_d$.
We can write down the $k$-step Bellman updates at the $i$-th iteration as follows, 
\begin{align}
    \label{equ:dpe_v_so}
    V^{(i)}(s_0 | o_d) \leftarrow
    \frac{R_d(\gamma^{k} - 1)}{k(\gamma - 1)}
    + \gamma^{k} V^{(i)}(s_{k}),
\end{align}
where the value function $V(s_k)$ in the RHS is updated using the standard value iteration, 
\begin{equation}\label{eq:dpe_v_s}
    V^{(i)}(s_0) \leftarrow \frac{R(\gamma^{k} - 1)}{k(\gamma - 1)} + \gamma^{k} V^{(i-1)}(s_{k}).
\end{equation}
Update \eqref{equ:dpe_v_so} is applied only when the transition involves a dispatch option. $R_d$ is the immediate reward from a dispatch option, i.e., the trip price. Update \eqref{eq:dpe_v_s} can be applied for both dispatch and reposition transitions, and
$R$ is the reward from the option, which is either the cost of idle movement ($\leq 0$) or the trip fee ($>0$).  The time discounting technique proposed in \cite{tang2019deep} is applied to the reward as in \eqref{equ:dpe_v_so}. The $\gamma$ is the discount factor between 0 and 1, and $k \geq 1$ is the time step index. 

The update equation \eqref{equ:dpe_v_so} is, in essence, similar to expected SARSA \cite{seijen2009}. The main difference is that expected SARSA uses empirical samples to approximate the expected value while DPE does so by directly learning a separate function approximator. To see that, note that $V(s)$ can be considered as the marginal expectation of the conditional value function $V(s|o)$, i.e., $V(s_k) = E_o(V(s_k | o))$. Hence the RHS of the updates in \eqref{equ:dpe_v_so} is directly using the expectation, i.e., $\frac{R_d(\gamma^{k} - 1)}{k(\gamma - 1)} + \gamma^{k} E_o(V(s_k | o))$ while expected SARSA uses empirical transition samples $\mathcal{S}_k$ starting from the state $s_k$ to approximate the corresponding expectation, i.e.,  $\frac{R_d(\gamma^{k} - 1)}{k(\gamma - 1)} + \frac{1}{|\mathcal{S}_k|}\gamma^{k}\sum_{o \in \mathcal{S}_k} V(s_k | o)$.
The overhead of learning two policies is minimal in this case since
both $V(s | o_d)$ and $V(s)$ are required for the value-based policy search as described in Section \ref{sec:planning}.

Similar to \cite{tang2019deep}, we use a neural network to represent the value function, but the training is different since we now maintain and update both the conditional value network $V(s | o)$ and the marginalized one $V(s)$.
We employ the same state representation and training techniques as introduced by \cite{tang2019deep}. Besides, for the conditional network we engage a separate embedding matrix to encode the option and use the multiplicative form to force interactions between the state features and the option embedding. Both $V(s | o)$ and $V(s)$ share the same state representation but have a separate branch for the output. An algorithm statement of DPE can be found in Appendix \ref{sec:app_dpe}.

\subsubsection{Dispatch Probabilities}\label{sec:dispatch_prob}
We estimate the dispatching probability $\pdk{s} \equiv \pi(o_d | s)$ by maximizing its log-likelihood on the transition data.
To generate the training data we collect drivers' historical trajectories including the descriptions of completed trips as well as the online and offline states. We use the states when the driver receives the trip request as the positive examples indicating the option $o$ being $o_d$.
For the negative examples we are unable to enumerate all possibilities considering the limited observed trajectories and the system complexity.
To that end we perform negative samplings. The negative examples we use for training are drivers' starting states of idle transaction in-between orders as well as the states when they become active or inactive. The training is done using one-month driver trajectories. Experiments on hold-out data sets show that the learned estimator achieves AUC of $0.867 \pm 0.010$ across multiple days and cities. Detailed results are presented in Appendix \ref{sec:app_dispatchprob}.

\subsection{Value-based Policy Search (VPS)}\label{sec:planning}
With the deep value network learned within the training framework in Section \ref{sec:vnet}, the optimal action is selected using decision-time planning when repositioning is triggered.  Specifically, we use our environment model to carry out planning of potentially multiple steps from the particular state $s_0$, the state of the agent when repositioning is triggered.  This allows us to evaluate at run-time the state-option pairs associated with $s_0$ and all the available repositioning options at that state and select the best move for the next step.  The next time repositioning is triggered, the same planning process is repeated.

We want to estimate $Q^*(s_0,o)$ associated with the optimal policy $\pir^*$ and the given $\pi_{d0}$, so that $o^* = \arg\max_o\Qhstar(s_0,o)$ gives the approximate optimal repositioning at decision-time with respect to a given dispatch policy.  
The one-step expansion per the environment model in Section \ref{sec:vnet} writes
\begin{equation}
	Q^*(s_0,o) = \xsupk{r}{0,1} + \xsupk{(V^*)}{t_{01}}(s_1),
\end{equation}
where $\xsupk{r}{0,1} \leq 0$ is the repositioning cost from $s_0$ to $s_1$, and $s_1$ is the state after repositioning $o$, with location $l_1$ and time $t_0 + t_{01}$.
$V^*$ is the state-value function associated with $\pir^*$ (and the given $\pi_{d0}$).  To make the time component of the input explicit, $\xsupk{(V^*)}{t_{01}}$ is the same value function with time component $t_{01}$ ahead of the decision-time $t_0$.  \emph{All value functions with a future time component are assumed to be properly discounted without clogging the notation in Section \ref{sec:planning}}. The discount factor 
is $\xsupk{\gamma}{t-t_0}$, where $t$ is discretized time component for the input state, and $t_0$ is the time for current decision point.
The duration that incurs cost $r$ is $\Delta t$, and the starting time for the cost is $t$.  The cost used in this paper is also properly discounted: $r \leftarrow \frac{\xsupk{\gamma}{t-t_0}r(\gamma^{\Delta t}-1)}{\Delta t(\gamma-1)}$.  

In practice, we use the state-value function $V$ learned in Section \ref{sec:vnet} to replace $V^*$ for computing $\Qhstar(s_0,o)$, writing $\xsupk{V}{t_{01}}(s_1)$ concisely as $\xsupk{V_1}{t_{01}}$. Then, 
\begin{equation}\label{eq:one-step}
	\Qhstar(s_0,o) = \xsupk{r}{0,1} + \xsupk{V_1}{t_{01}}.
\end{equation}  
That is, that the one-step expansion renders a greedy policy by selecting the repositioning movement leading to the next-state with the highest value given by the state value networks, $V$.  We also notice that $\Qhstar(s_0,o)$ is in fact $Q^{\pi_0}(s_0,o)$ in this case because $V$ is computed by policy evaluation on historical data generated by $\pi_0$.  Hence, finding the optimal option to execute by $o^* = \arg\max_o\Qhstar(s_0,o)$ is essentially one-step policy improvement in generalized policy iterations.

We can use our environment model to expand $\Qhstar$ further:
\begin{equation}
	\Qhstar(s_0,o) = \xsupk{r}{0,1} + \pdk{1}\Vhstar(s_1 | dispatch) + \pidk{1}\Vhstar(s_1 | idle).
\end{equation}
$\pdk{1}$ is the dispatch probability at $s_1$, and we use the conditional value networks discussed in Section \ref{sec:vnet}, denoted by $\xsupk{\tilde{V}}{t_{01}}_1$, for $\Vhstar(s_1 | dispatch)$.  When the driver is idle, the immediate next option has to be a reposition, so we have $\Vhstar(s_1 | idle) = \max_j \Qhstar(s_1,o_j)$, where $o_j$ is the second-step reposition option.  $\Qhstar(s_0,o)$ can be recursively expanded, eventually written in terms of the given estimated state-value function.

\begin{figure}
\begin{center}
        \includegraphics[width=0.7\linewidth]{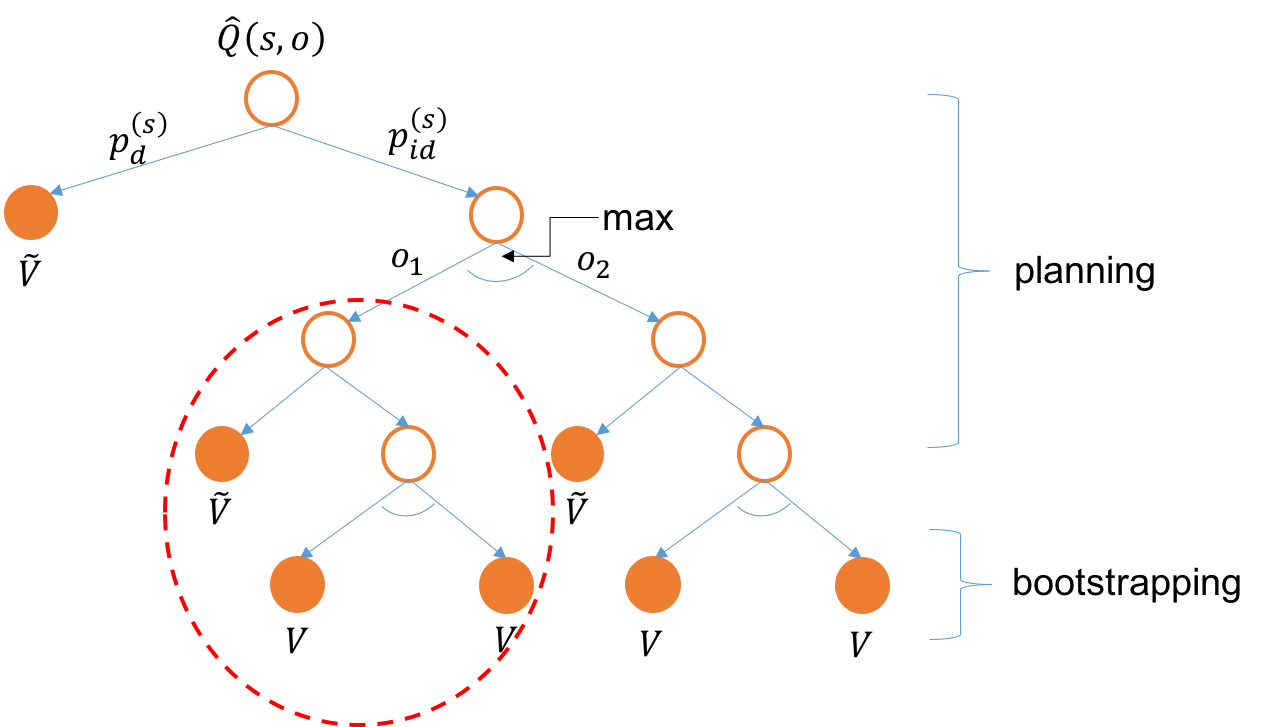}
        
	\caption{Planning + bootstrapping perspective of the proposed algorithm.  At the leaf nodes, bootstrapping is used through the given state-value function $V$.  The subtree within the red loop illustrates the back-up of values.}
	\label{fig:planning_bootstrapping}
\end{center}
\end{figure}

A two-step expansion is that
\begin{align}
	\Qhstar(s_0,o) &= \xsupk{r}{0,1} + \pdk{1}\xsupk{\tilde{V}}{t_{01}}_1 + \pidk{1}\max_j \Qhstar(s_1,o_j) \nonumber \\
			   &= \xsupk{r}{0,1} + \pdk{1}\xsupk{\tilde{V}}{t_{01}}_1 + \pidk{1}\max\left\{ \max_{j\neq 1}\xsupk{r}{1,j} + \xsupk{V_j}{t_{0j}}, \xsupk{V_1}{t_{01}+L}\right\}. \label{eq:two-step}
\end{align}
For a three-step expansion, we further write
\begin{align}\label{eq:three-step}
	\Qhstar(s_1,o_j) &= \xsupk{r}{1,j} + \pdk{j}\xsupk{\tilde{V}_j}{t_{0j}} + \pidk{j}\max\left\{ \max_{k\neq j}\xsupk{r}{j,k} + \xsupk{V_k}{t_{0k}}, \xsupk{V_j}{t_{0j}+L} \right\}.
\end{align}
In the above equations, $t_{0j} := t_{01} + t_{1j}, t_{0k} := t_{01} + t_{1j} + t_{jk}$, are the total ETA of two-step and three-step repositions respectively.  
Figure \ref{fig:planning_bootstrapping} summarizes the process of our value-based policy search.

We can interpret the above decision-time planning through path value approximation.
For repositioning, not only the destination cell is important, the route is important as well, since the driver may
be matched to orders along the way, generating income.
Our approach can be viewed as computing the long-term expected values of $n$-step look-ahead repositioning paths from the current spatiotemporal state, selecting the optimal one, and then executing its first step.  
An $n$-step look-ahead path also corresponds to a path from the root to a leaf node with a depth of $2n$ in Figure \ref{fig:planning_bootstrapping}. We show in Section \ref{sec:impl} that selecting the max-value path is equivalent to $o^* = \arg\max_o \Qhstar(s_0,o)$ in terms of the action executed. 
The new data generated by the current learned policy is collected over time and can be used to update the state-value function, which in turn updates the policy through decision-time planning.

\subsubsection{Practical Implementation}\label{sec:impl}
Our implementation of the policy search algorithm (VPS) is based on the path-value perspective instead of directly implementing \eqref{eq:one-step} to \eqref{eq:three-step}.  The reason is that the latter does not allow batch inference of the state value networks, which is crucial for fast computation.  We first show the equivalence of the two approaches taking the three-step expansion as an example.

By convention, for the case of staying at the same location ($j=k$), $\xsupk{r}{j,k} = 0$ and $t_{0k} = t_{0j}+L$.  Then, we have
$\max\left\{ \max_{k\neq j}\left\{\xsupk{r}{j,k} + \xsupk{V_k}{t_{0k}}\right\}, \xsupk{V_j}{t_{0j+1}} \right\} = \max_{k}\left\{\xsupk{r}{j,k} + \xsupk{V_k}{t_{0k}}\right\}.$
Through further transformation, we get $\Qhstar(s,o)$
\begin{align}
&= \xsupk{r}{0,1} + \pdk{1}\tilde{V}(s_1) + \pidk{1}(\max_j \Qhstar(s_1,o_j)) \nonumber\\
&= \xsupk{r}{0,1} + \pdk{1}\tilde{V}(s_1) + \pidk{1}\max_j \left\{ 
 \xsupk{r}{1,j} + \pdk{j}\tilde{V}(s_j)
 + \pidk{j}\max_{k}\left\{\xsupk{r}{j,k} + \xsupk{V_k}{t_{0k}}\right\}\right\}\nonumber\\
&= \xsupk{r}{0,1} + \pdk{1}\tilde{V}(s_1)  + \pidk{1}\max_{j,k} \left\{
\xsupk{r}{1,j} + \pdk{j}\tilde{V}(s_j)
 + \pidk{j}(\xsupk{r}{j,k} + \xsupk{V_k}{t_{0k}})\right\} \nonumber\\
&= \max_{j,k}\Big\{\xsupk{r}{0,1} + \pdk{1}\tilde{V}(s_1) +\pidk{1}(\xsupk{r}{1,j} + \pdk{j}\tilde{V}(s_j)
 + \pidk{j}(\xsupk{r}{j,k} + \xsupk{V_k}{t_{0k}}))\Big\}. \label{eq:algo}
\end{align}
Equation \eqref{eq:algo} shows the correctness of our implementation, supported by the following proposition:

\textit{The state-option value that we use for decision-time planning (Eq \eqref{eq:three-step}) is the maximum expected value of the three-step paths with $o$ being the first step.}

We have developed a highly efficient implementation of VPS that has met the strict production requirement on latency for our real-world experiment. Since the vehicle repositioning requests can be processed independently, the proposed algorithm is horizontally scalable.

Our algorithm based on \eqref{eq:algo} is split into two phases. First, we generate all the paths of a certain length originating at the current grid cell using breadth-first-search.
Second, we calculate the value of each path and select the first step of the path which has the maximum value as our repositioning action.
Keeping two separate phases has the benefit of allowing batch inference of the state value networks model in the second phase. We will now describe the details of each phase.

\paragraph{Path generation}
Each city is divided into hexagon cells, each covering an equal size area. In each grid cell, we use the pick-up points (from an in-house map service) as the candidate repositioning destinations. Some of those cells might be ``inaccessible"(e.g. lakes, rivers, mountains) or might be missing well-known pick-up points (e.g. low population density areas). These cells are excluded from our search.

The main parameter for the path generation is the maximum path length that our algorithm should consider. We call this parameter the \emph{expansion depth} (also from the perspective of \eqref{eq:two-step} and \eqref{eq:three-step}).
Our algorithm then uses a breadth-first search (BFS) to generate paths of a maximal length no longer than the expansion depth. 
Because of the presence of invalid cells, it may happen that no generated path reaches the search depth. In the infrequent case where no cell is valid within the search depth distance to the origin cell, we expand our breadth-first search to the city limit to find the closest valid cells to the current cell and use those as repositioning candidates.  
We also note that with bootstrapping the state values, the path values account for an horizon far beyond the expansion depth. 

\paragraph{Step selection} Once a set of paths has been generated, depending on the length reached, our algorithm applies a different formula to each path in order to calculate its value, e.g., \eqref{eq:algo} being applied to the paths of length three, and \eqref{eq:one-step} being applied to  unit-length paths.
Finally, the algorithm returns as reposition action the first step of the path which has the maximum value.

\paragraph{Long search} Occasionally, it can take the above search algorithm many repositioning steps to guide a driver to an area of higher value because e.g., the driver is stuck in a large rural area with low demand, leading to low repositioning efficiency.
To tackle this issue, we have developed a special mechanism called long-search, triggered by idling for more than a time threshold (e.g. 100 minutes). Long-search repositions drivers to a globally relevant area in order to help them escape from `local minima', by choosing among destinations with globally top state values  (with coarser time window) discounted by their respective travel times. Specifically, for each hex cell, we use the most popular pick-up location (provided by an in-house map service) to represent it and consider the location?s state value given by the spatiotemporal value network developed in Section \ref{sec:vnet}. 
Then, we retrieve the locations corresponding to the top 200 state values averaged over 20-minute intervals. The values are pre-computed and stored in a look-up table.
During long search, the destination with top time-discounted state value for that hour with respect to the current location of the vehicle is selected for reposition. Time discounting is done by considering the estimated travel time $\tau$ (in minutes) from the current location of the vehicle to the candidate destination and using a multiplicative discount factor $\lambda^{\tau/10}$, where $\lambda$ is the discount parameter in Section \ref{sec:planning}.

\paragraph{Computational complexity \& approximation error}  
The complexity of BFS on the grid cell graph (each unit being a hexagon cell) for solving \eqref{eq:three-step} is $O(n^2)$ in the expansion depth of $n$. As in Section \ref{sec:vps-depth}, we have found that a small $n$ works the best in practice, which limits the impact of this quadratic behavior.  Larger depth values would result in accumulation of errors introduced by model assumption and estimation, which outweighs the benefits of the models. The primary source of approximation error in planning is the estimation of dispatch probabilities. Per \eqref{eq:two-step} and \eqref{eq:three-step}, the multiplicative factors for the value terms $V$'s and $\tilde{V}$'s are products of the dispatch probabilities at each expansion level above them. The estimation error is hence compounding in nature, and the predictive capability of the product of dispatch probabilities as a binary classifier quickly deteriorates as $n$ increases, considering the AUC of the dispatch probability model.

\section{Learning Action-values for a Large Fleet}\label{sec:action-val}
There are more challenges for managing the repositioning of a large fleet. More global coordination is required so that repositioning does not create additional supply-demand imbalance. As a simple example, under independent individual strategy, all the vehicles with the same spatiotemporal state may be repositioned to the same high-value destination. This `over-reaction' phenomenon is undesirable because it may cause the seemingly high-value destination to become over-supplied in the future while leaving the other areas under-supplied. Hence, it is necessary to make modification to the learning algorithm so that coordination among a group of vehicles can be better induced into the repositioning policy.

Similar to the small-fleet case, we learn a single shared action-value network to generate policies for each managed vehicle. There are multiple benefits associated with this decision. First of all, the number of agents in the vehicle repositioning problem changes over time, since drivers may get online and offline at different times, and existing vehicles may exit and new vehicles enter the system, making it hard to manage agent-specific networks especially for new agents without any data. A single value network is agnostic to the number of agents in the environment, thus naturally handling this issue well. In contrast, learning agent-specific value functions, as in many multi-agent reinforcement learning (MARL) methods, is challenging to fit in this setting. The second advantage is the ability to crowd-source training data. Abundance and diversity in experience samples are especially important in offline RL. Training a separate network for each agent may not be the most effective way to utilize the data because the data specific for each agent is limited and may cover only a small part of the state-action space. Lastly, from a practical perspective, a single value network requires a simpler training pipeline, making it deployment-friendly for a high usage service like vehicle repositioning.

We choose to learn the action-value function $Q(s,o)$ directly in this case. The motivations mostly come from the ease of incorporating real-time contextual features and algorithmic elements of MARL, which we will elaborate in the subsections below. Incorporating contextual features in VPS is harder, because that requires prediction of those features for future times (see \eqref{eq:one-step},\eqref{eq:two-step}, and \eqref{eq:three-step}), introducing more sources of error. Another crucial reason for this algorithmic choice is speed. Our method is designed to be production-ready, which means that the online planning component has to meet strict latency requirements. An action-value network allows for much faster decision-time planning than VPS because no tree search is  required and batch inference is possible. For a concrete comparison, the QPS (queries-per-second) of a single action-value network is about six times higher than that of VPS, which is a difference of go or no-go in the case of managing a large fleet. 

Specifically, we use deep SARSA to learn the state-action value function using a similar training framework as CVNet \cite{tang2019deep}. The structure of the Q-network consists of three blocks of layers: the embedding layer, the attention layer, and the output layer. The architecture is illustrated in Figure \ref{fig:sarsa_arch}. We use the same state representation for spatial information through hierarchical sparse coding and cerebellar embedding as in CVNet. The destination of the action selected by the policy is determined among a set of predefined pick-up points, and the policy is also supplemented by Long Search as discussed in Section \ref{sec:impl}.

It is easy to see that vanilla deep SARSA in this case is indeed just VPS with an expansion depth of 1. 
Expanding $Q(s,o)$ and assuming a stochastic policy $\pi$, we have
\begin{eqnarray}
    Q^\pi(s,o) &=& E_{ o'\sim\pi(\cdot|s')}\left[r(s,o) + \gamma Q^\pi(s',o')\right] \\
            &=& r + \gamma V^\pi(s'),
\end{eqnarray}
where $s'$ is the next state after $s$ with the option $o$. Relating to \eqref{eq:one-step}, the connection is now established. Albeit its simpler form, deep SARSA allows for a more natural incorporation of several highly desirable algorithmic elements as described below.

\subsection{Supply-demand Context}\label{sec:sd-context}
\begin{figure}
\begin{center}
        \includegraphics[width=1.0\linewidth]{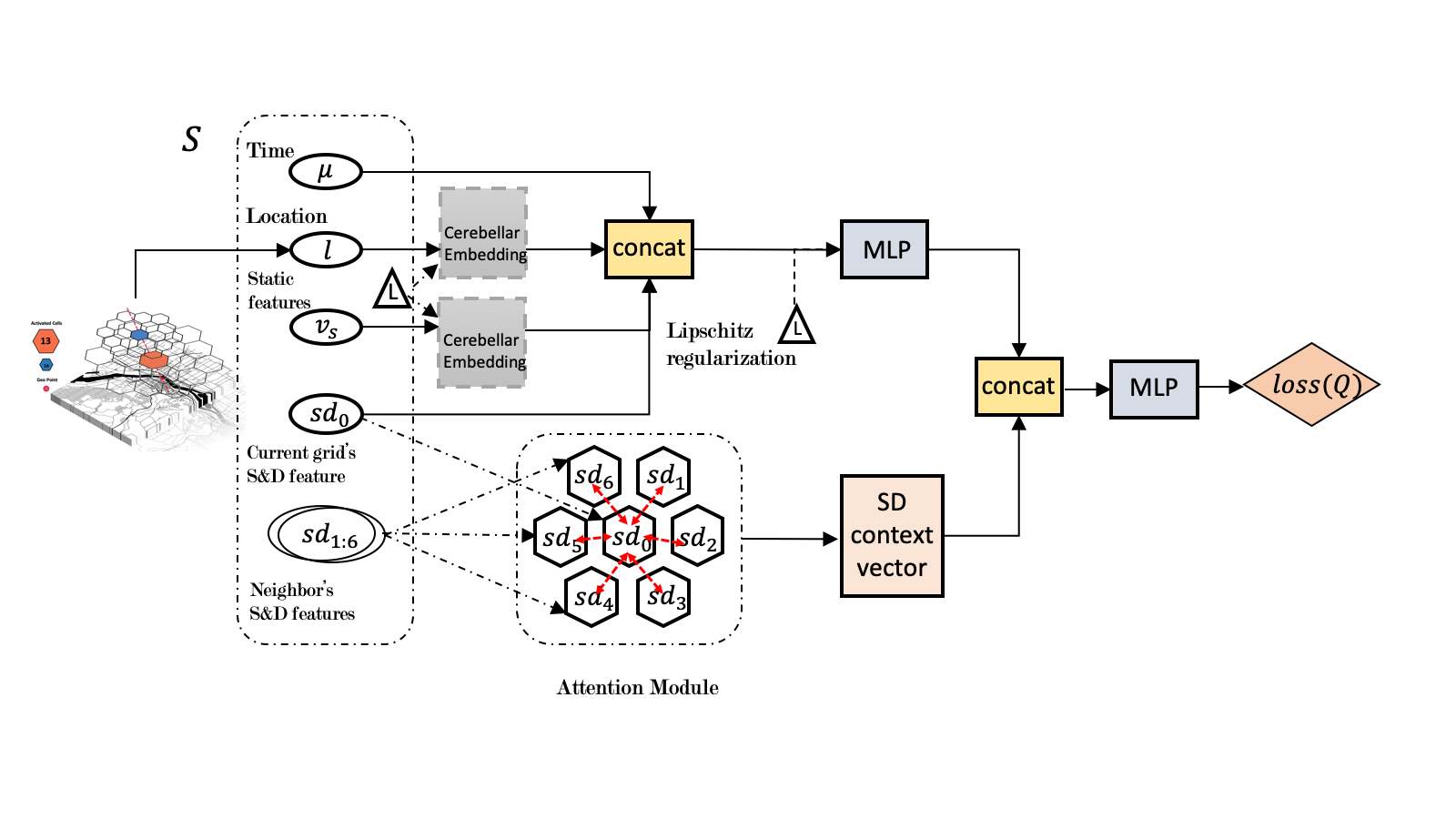}
	\caption{The Q-network architecture for SARSA.}
	\label{fig:sarsa_arch}
\end{center}
\end{figure}

To enable the policy of the vehicles to better induce coordination and to be more adaptive to the dynamic nature of the environment, we augment the state space with additional supply-demand (SD) contextual features in the neighborhood of the agent. Typical SD features are the numbers of open orders and idle drivers to be matched. These help characterize the state of the vehicle and its surrounding environment more accurately, allowing for better state representation and responsiveness to changes in the environment. For this purpose, we include the SD features of six neighboring hexagon grid cells as well as that of the current cell. These quantities are local and are specific to each hex cell. The SD features of the vehicle?s state includes SD quantities of seven hex cells, six neighboring cells and the current cell. For each cell, the vector representation of the SD features include real-time numerical values such as number of idle drivers, number of total requests, and number of orders that have not been assigned yet within the geo-boundary of the cell. Those quantities are aggregate values computed over a 10-min sliding window.
This structure design is intuitive because the action space of the agent covers destinations in the same set of seven grid cells. Specifically, we adopt the general form of global attention mechanism proposed in \cite{luong2015effective}. As illustrated in Figure \ref{fig:sarsa_arch}, the attention layer assigns scores to each pair of SD features (current cell and one of its neighbor cell)
\begin{equation}\label{eq:attention}
   \alpha_{i}=\mbox{ softmax}(sd_{0}^\top W_{\alpha} sd_{i})
\end{equation}
where $i \in \mathbb{Z}, i=[1..6]$, and $ W_{\alpha}$ is a trainable weight matrix in the attention layer. The scores are then used to re-weight the neighboring SD vectors, obtaining a dense and robust context representation vector. The motivation of constructing such an attention module in our learning network is to cast more weights into nearby grids possessing better supply-demand ratio than the current cell. The mechanism utilizes real-time contextual features to model the relationships between spatial regions according to the change of the dynamic. For nearby cells sharing similar spatial and static features, more attention would be given to the action destination with abundant ride requests.

\subsection{Stochastic Policy}\label{sec:stoch-policy}
With the deterministic policy in Section \ref{sec:vps}, vehicles in the same state will be repositioned to exactly the same destination. When the algorithmically controlled fleet is small compared to the general vehicle pool (and assuming that the size of the environment is inline with that of the general pool), the vehicles in the fleet can be essentially treated independently because the probability of multiple vehicles being in the same state is small. As the size of the fleet increases, it happens more often that the vehicles would come across each other, and the effect of the `over-reaction' phenomenon becomes more severe. To mitigate this undesirable effect, we randomize the reposition action output by adding a softmax layer to obtain the Boltzmann distribution
\begin{equation}\label{eq:softmax}
    \sigma(\bold{q})_k = \frac{\exp(q_k)}{\sum_j\exp(q_j)}, \forall k\in\mathcal{K},
\end{equation}
where $\bold{q}$ represents the vector of reposition action-values output, and $\mathcal{K}$ is the set of eligible destinations. While this is a common approach in reinforcement learning to convert action-values into action probabilities for a stochastic policy, we argue that it is particularly appealing in the case of vehicle repositioning: First, the sign of the action values would not be a concern. As we will see in Section \ref{sec:sd-reg}, sometimes the action values are penalized to account for the current SD condition, making negative resulting values possible. Second, we dispatch the vehicles following the action distribution $\sigma(\bold{q})$, so when there are multiple idle vehicles in the same area at a given time, we basically dispatch them in proportion to the exponentiated values, which aligns with intuition. It is easy to see that this transformation can be applied to VPS in general to derive a stochastic policy.
In Section \ref{sec:sim-stoch} of the simulation experiments, we demonstrate the importance of stochastic policy in achieving better performance when managing a large fleet.

\subsection{Decision-time SD Regularization}\label{sec:sd-reg}
The semi-MDP formulation in Section \ref{subsec:formulation} is from a single vehicle's perspective, and the action value does not take into account the supply-demand gap at the destination explicitly (but through part of the state features). 
To explicitly regularize the action values, and hence the action distribution, so that the system generates SD-aware repositioning actions, we penalize the action values by their respective destination SD gaps in a linear form,
\begin{equation}
    q^\prime_k := q_k + \alpha g_k, \forall k \in \mathcal{K},
\end{equation}
where $q^\prime_k$ is the penalized version of the Q-value $q_k$ in \eqref{eq:softmax}, and $g_k$ is the SD gap in the destination cell $k$.
Considering the construction of a stochastic policy through the Boltzmann distribution in \eqref{eq:softmax}, the SD-gap penalty is in fact multiplicative on the action distribution. We have empirically observed that it helps to truncate the penalty by setting a threshold on the SD gap, i.e.,
\begin{equation}
    q^\prime_k := q_k + \alpha g_k\bold{1}_{(g_k > \beta)}, \forall k \in \mathcal{K},
\end{equation}
where $\beta$ is the threshold parameter for SD gaps and is usually city-specific.

To better understand what we are doing here, it helps to relate to an assignment formulation which assigns idle vehicles from every grid cell at the decision time to destination cells that are under-supplied while maximizing some utility function, e.g., the action-value function. SD regularization with Boltzmann action distribution can be thought as soft-constrained version of the above decision-time planning. A major advantage over the assignment formulation is that it is generally less sensitive to perturbation in the input SD data, which is dynamic and prone to prediction errors.

\section{Simulation Experiments}
We describe our simulation experiments in this section and discuss empirical results and observations.
We have developed a realistic simulation environment for evaluating vehicle repositioning policies based on prior works on order dispatching for ride-hailing platforms \cite{tang2019deep,xu2018large}.  This environment captures all the essential elements of a large-scale ride-hailing system. We emphasize that our proposed algorithms do not require the simulation environment for training.
For realistic modeling of the real-world MoD system, we have augmented the multi-driver simulator in \cite{tang2019deep} with individual vehicle repositioning capability.  
In this environment, one can specify a given number of vehicles to follow a particular repositioning policy $\pir$.  
We set this number to ten for the experiments in Sections \ref{sec:vps-depth} and \ref{sec:benchmarking}, and it is set to 1600 for Section \ref{sec:sim-stoch} and 3200 for Sections \ref{sec:sim_context} and \ref{sec:sim_sd_reg}.
A fixed order dispatching policy $\pi_d$ assigns trip orders to the drivers in batch windows, and minimizes the within-batch total pick-up distance. In addition, order dispatching can interrupt a repositioning action.  The same evaluation environment framework has been deployed for hosting KDD Cup 2020 RL Track \cite{qin2020kddcup}. More details on the simulation environment and the data sets used can be found in Appendix \ref{sec:app_sim}.

The experiments are divided into two types, individual-oriented and group-oriented, for testing the performance of VPS and deep SARSA using individual- and group-level metrics respectively. As defined in Section \ref{subsec:formulation}, the individual-level IPH for a driver $x$ and the group-level IPH for a driver group $X$ are \eqref{eq:ind_iph} and \eqref{eq:agg_iph} respectively. For the purpose of the experiments below, both the income and online hours are accumulated over the course of the specific tests. For all the experiments, We report group-level IPH. In the small-fleet cases, where all the managed vehicles' online hours are the same in the simulations, the group-level IPH is equivalent to the mean of the individual-level IPHs.

\subsection{Expansion Depth of VPS}\label{sec:vps-depth}
\begin{figure}
\begin{center}
    \begin{subfigure}{.5\textwidth}
    \includegraphics[width=1.0 \linewidth]{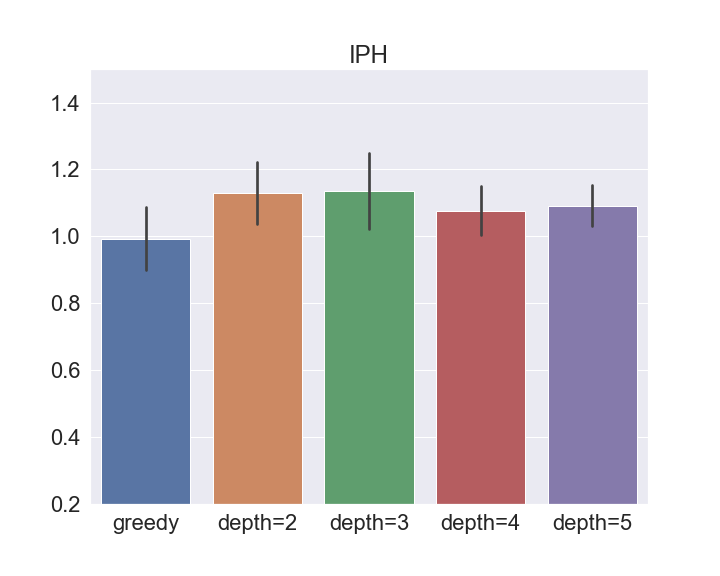}
    \caption{IPH v.s. expansion depth. All the results have been standardized w.r.t. that of greedy, i.e., depth=1.}
    \label{fig:vps_depth_iph}
    \end{subfigure}
    \begin{subfigure}{.5\textwidth}
    \includegraphics[width=1.0 \linewidth]{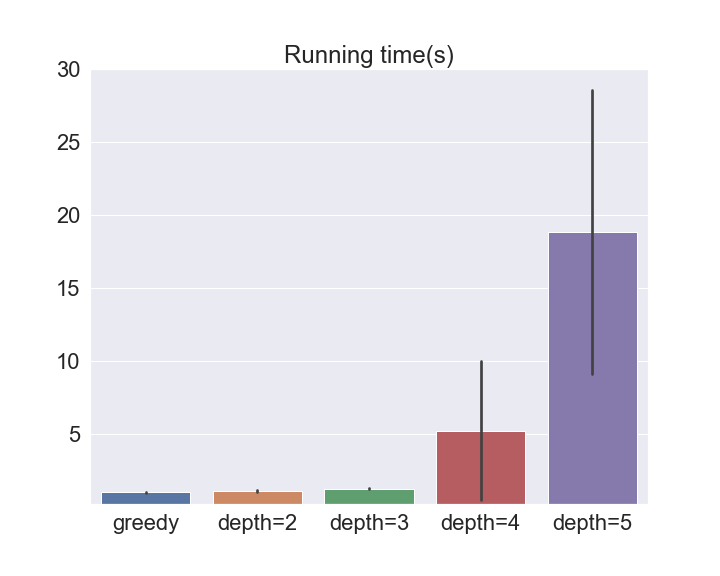}
    \caption{CPU time per simulation (standardized w.r.t. depth=1) v.s. expansion depth}
    \label{fig:vps_depth_cpu}
    \end{subfigure}
	\caption{VPS expansion depth selection (10 drivers).}
	\label{fig:depth_sel}
\end{center}
\end{figure}

We benchmark implementations of VPS with expansion depth from 1 (Greedy) to 5 through simulation with data from a different city than the three cities in Section \ref{sec:benchmarking}.  Each method was tested using 10 random seeds, and we report both mean and standard deviation of the income rates (IPH) computed over three different days. 
To understand the effect of the expansion depth in VPS, we see that it controls the extent to which we leverage the estimated environment model information to compute the approximate action-values (see Figure \ref{fig:planning_bootstrapping}). The hope is that by exploiting some model information of the MDP, we would be able to learn better action-values than a completely data-driven model-free method. On the other hand, the errors in model estimation, i.e., dispatch probability model, are compounded quickly as the expansion depth increases. So a deeper expansion will likely hurt the quality of the learned action-values. Empirically, this is indeed observed. From \ref{fig:vps_depth_iph}, we see that all the expansion depths greater than 1 yield higher average IPH  than an expansion depth of 1, demonstrating that the policy search with model information does help improve the policy. In particular, the depths of 2 and 3 lead to more than 10\% higher average IPH than the greedy method, and they also perform significantly better than the depths of 4 and 5, where the error in model estimation has apparently kicked in. Figure \ref{fig:vps_depth_cpu} shows the rapid growth in CPU time per simulation with respect to the expansion depth. The computation corresponding to a depth of more than 3 is unacceptably expensive. The mean IPH's for the depth of 2 and 3 are basically the same, but the standard deviation for the depth of 2 is noticeably smaller than that of 3, and the policy search with a depth of 2 is also faster.
According to Figure \ref{fig:depth_sel}, we have chosen an expansion depth of 2 for both the simulation and the real-world experiments based on average performance, robustness, and computational efficiency. 
It should be stressed that the expansion depth here is a different concept from the optimization horizon of the MDP. Regardless the value of the expansion depth, the state values that we use for bootstrapping always have the same horizon, i.e., infinite horizon.

\subsection{Benchmarking VPS}\label{sec:benchmarking}
We benchmarked our algorithm in the simulation environment generated by ride-hailing trip data from three mid-sized cities, which we denote by A, B, and C. Table \ref{tab:small_fleet_cities} shows the basic background information of the cities. Ten drivers are selected for each experiment out of hundreds of drivers in the environment. We chose three different days, and for each city and each day, we use five random seeds, which give different initial states for the experiment.
We compare four algorithms:
\emph{Random}: This policy randomly selects a grid cell from the six neighbouring cells and the current location as the repositioning destination.
\emph{Greedy}: This is the implementation of \eqref{eq:one-step}, which is one-step policy improvement of the generalized policy iterations.  
\emph{MAB}: This is a multi-arm bandits algorithm adapted from \cite{garg2018route}, with 7 arms corresponding to the 7 possible actions and an arm's value being the average return for repositioning from the current location to the destination specified by the arm. The policy is generated by UCB1 \cite{auer2002finite} with one day of pretraining. 
\emph{VPS}: This is the value-based policy search algorithm described in Section \ref{sec:planning}.  An expansion depth of two is chosen by benchmarking different expansion depths as in Figure \ref{fig:depth_sel}.

\begin{figure}
\begin{center}
        \includegraphics[width=0.9\linewidth]{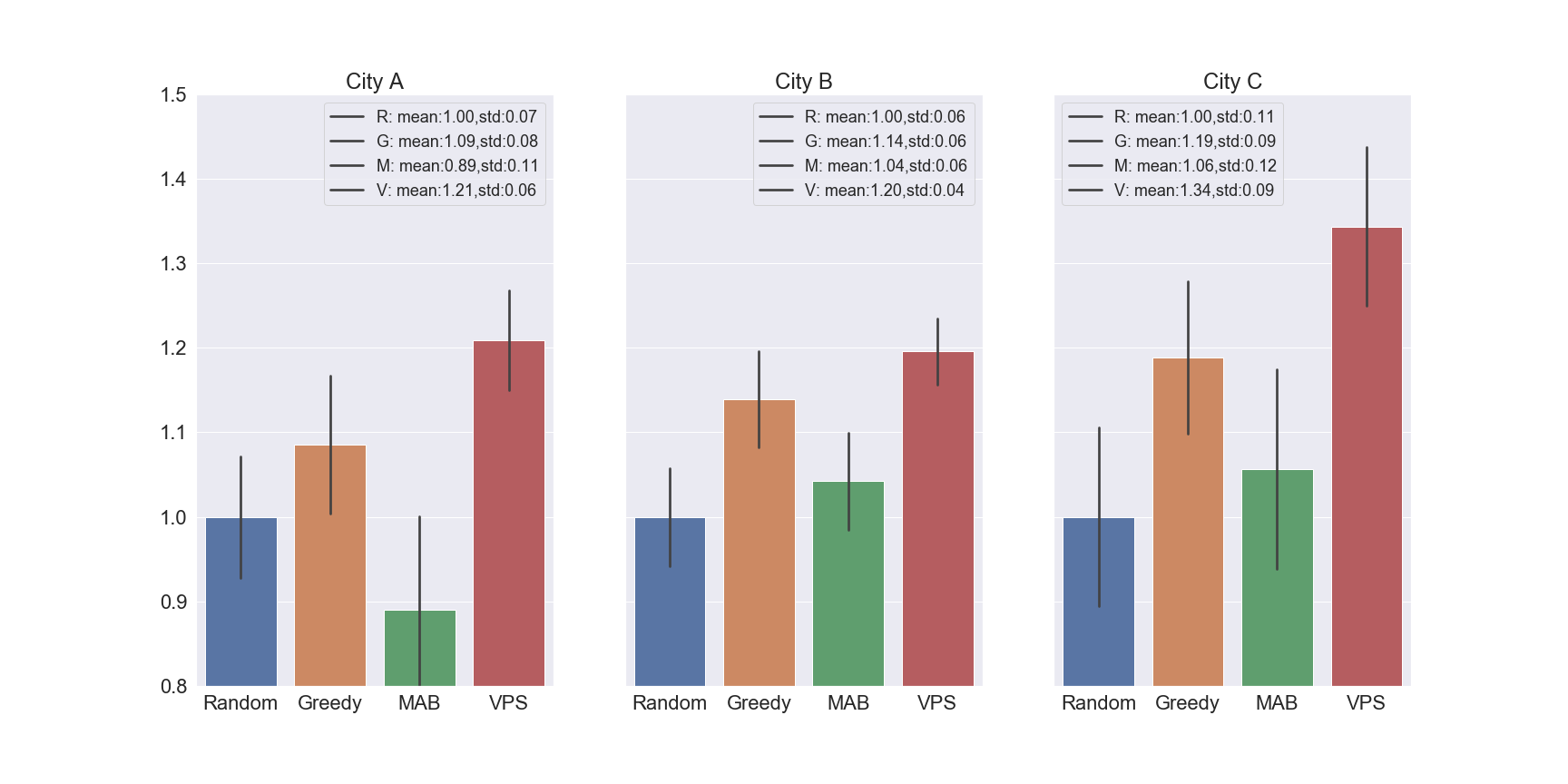}
	\caption{Simulation experiment results comparing income rates of Random, MAB, Greedy (Baseline) and VPS policy using the multi-driver simulation environment. Results are normalized with respect to the mean Random.}
	\label{fig:sim_combined}
\end{center}
\end{figure}

We computed the average income rate across three days for each random seed. We then compare for the three methods the mean and standard deviation of the five random seeds results. We can see from Figure \ref{fig:sim_combined} that both VPS and Greedy methods consistently outperform Random and MAB across the three cities.  Moreover, VPS yields significantly higher income rates than Greedy, with 12\%, 6\%, and 15\% improvement in the mean for each of the three cities respectively. The comparison between VPS, Greedy and MAB shows the advantage of a longer horizon in the setting of a long-term objective.

\subsection{Stochastic Policy}\label{sec:sim-stoch}
\begin{figure}
\begin{center}
        \includegraphics[width=0.8\linewidth]{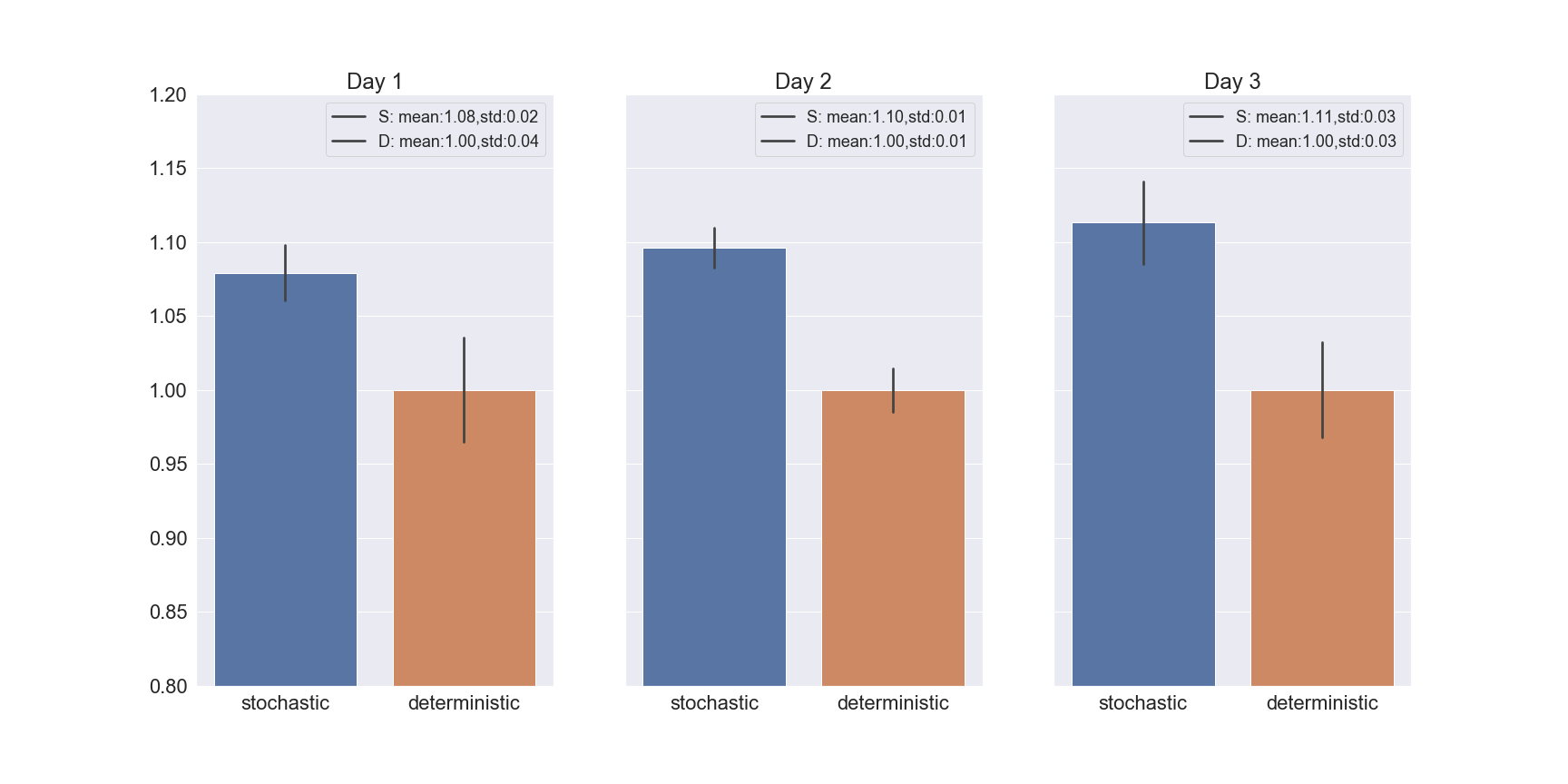}
	\caption{IPH comparison between VPS with stochastic and deterministic policies for City K. Results are standardized with respect to the numbers for the deterministic policy.}
	\label{fig:vps_stocastic_determinsitic}
\end{center}
\end{figure}

To demonstrate the necessity of a stochastic policy for managing a large fleet, we create a new version of VPS with stochastic policy, VPS-stoch, by applying the softmax function to the approximate action values $\Qhstar(s_0,o)$ as in Section \ref{sec:stoch-policy}. Both VPS and VPS-stoch have the same expansion depth and other parameters. The evaluation environment is generated using data from city K, which is one of the cities for the large-fleet real-world deployment. We set the number of algorithmically-guided vehicles in the environment to 1600. Figure \ref{fig:vps_stocastic_determinsitic} compares the performance in IPH of VPS and VPS-stoch. It is clear that VPS-stoch consistently outperforms VPS with a deterministic policy by about 10\% across different days in the large-fleet setting. We expect the performance comparison between deep SARSA with greedy policy and the one in Section \ref{sec:action-val} to be similar.

\subsection{SD Contextual State Features}\label{sec:sim_context}
We carry out ablation study to justify the use of SD contextual state features in deep SARSA. To make sure that the contextual features in the training data aligns with the simulation environment, we run a deep SARSA agent trained on historical data in the simulation environment to collect six weeks of training data. We then train two new deep SARSA agents with and without the contextual features and evaluate them on a set of different dates. In Figure \ref{fig:sd_nosd_reg.png}, we present the comparison results for the time interval of 3-5pm, which we have found most significant within the experiment period (11am - 7pm). We surmise that such an observation is because the SD conditions during lunch hours and evening rush hours are relatively stable, whereas that during the off-peak hours is more volatile, making the SD contextual features more useful there.
\begin{figure}
\begin{center}
        \includegraphics[width=0.8\linewidth]{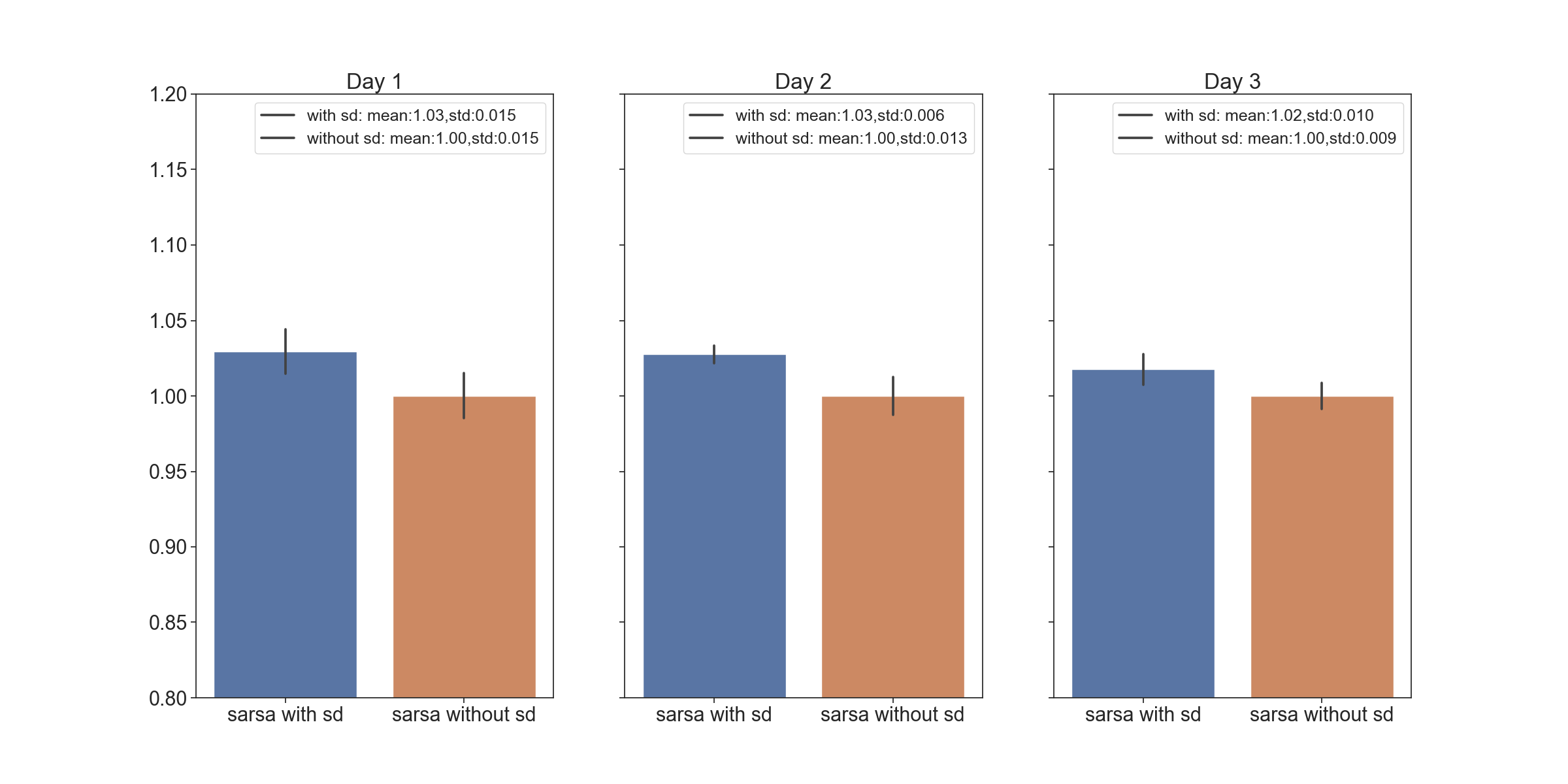}
	\caption{IPH comparison between the versions of SARSA with and without SD features. Results are standardized with respect to the numbers for SARSA without SD features. The results are for the time interval 3-5pm over three different days.}
	\label{fig:sd_nosd.png}
\end{center}
\end{figure}

\subsection{SD Regularization}\label{sec:sim_sd_reg}
\begin{figure}
\begin{center}
        \includegraphics[width=0.8\linewidth]{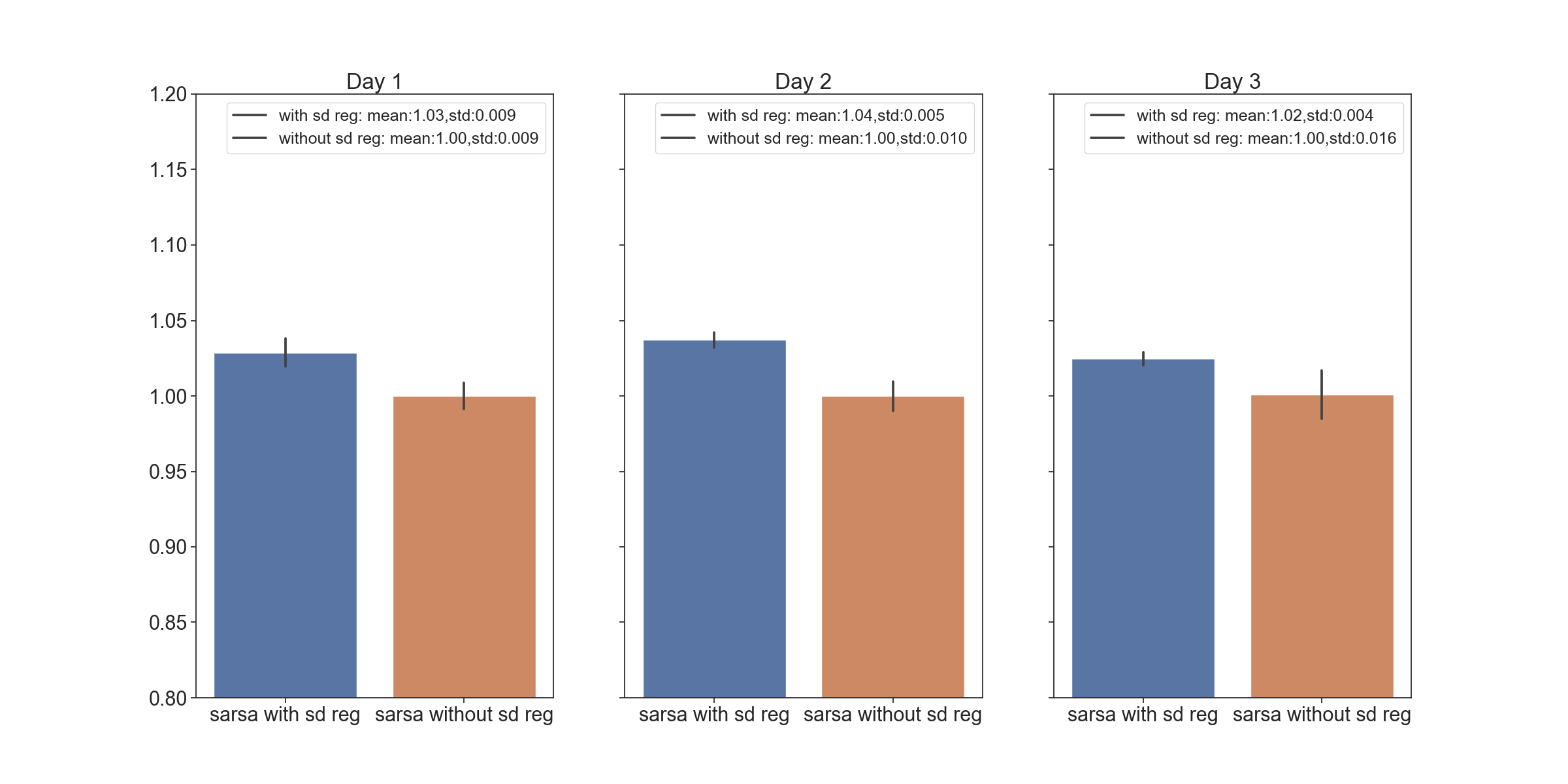}
	\caption{IPH comparison between the versions of SARSA with and without SD regularization. Results are standardized with respect to the numbers for SARSA without SD regularization. The results are for the first hour of the experiment period with all the drivers online (no new drivers getting online or existing drivers getting offline) in the environment.}
	\label{fig:sd_nosd_reg.png}
\end{center}
\end{figure}

To demonstrate the effect of SD regularization on the action values, we present ablation study results comparing deep SARSA with and without SD regularization at decision time. As one could see from figure \ref{fig:sd_nosd_reg.png}, the IPH of SARSA with SD regularization is significantly higer than SARSA without SD regularization across three different days, verifying that it is important to consider the locations of other idle vehicles for repositioning a large fleet. 

\begin{figure}
\begin{center}
        \includegraphics[width=0.8\linewidth]{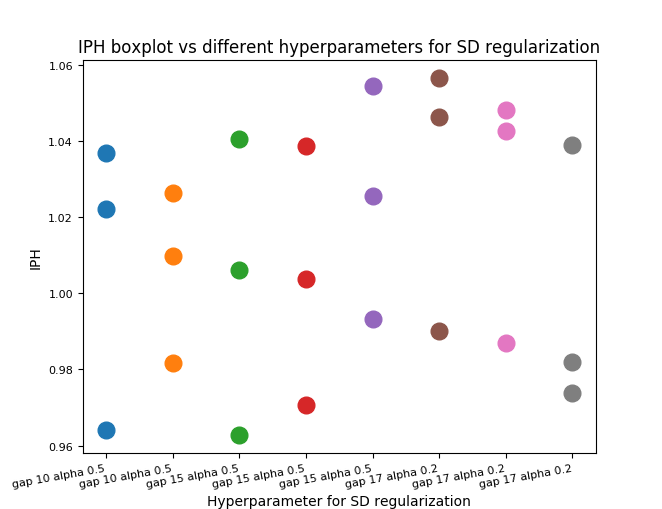}
	\caption{IPH comparison between different parameters for SD regularization (SD gap threshold and penalty parameter $\alpha$ in Section \ref{sec:sd-reg}). The data points are the IPH values resulted from three random seeds for each combination of the parameters on the x-axis. The data points with the same color correspond to the same parameter combination.}
	\label{fig:sd_params.png}
\end{center}
\end{figure}

We also present in Figure \ref{fig:sd_params.png} the results from using different combinations of the SD regularization parameters, i.e., the SD-gap threshold $\beta$ and the penalty parameter $\alpha$. 
Figure \ref{fig:sd_params.png} plots the IPH values resulted from three random seeds for each combination of the parameters (gap and alpha) on the x-axis. The data points with the same color correspond to the same parameter combination. This plot shows that the SD regularization parameters have significant impact on the IPH performance, with the empirically best combination at $\beta=17$ and $\alpha=0.2$.
Within the respective appropriate range, the general observation is that larger penalty and threshold are preferred possibly because smaller thresholds tend to introduce noise from the dynamic changes in the SD context and smaller penalties diminish the effect of regularization.

\begin{figure}
\begin{center}
        \includegraphics[width=0.8\linewidth]{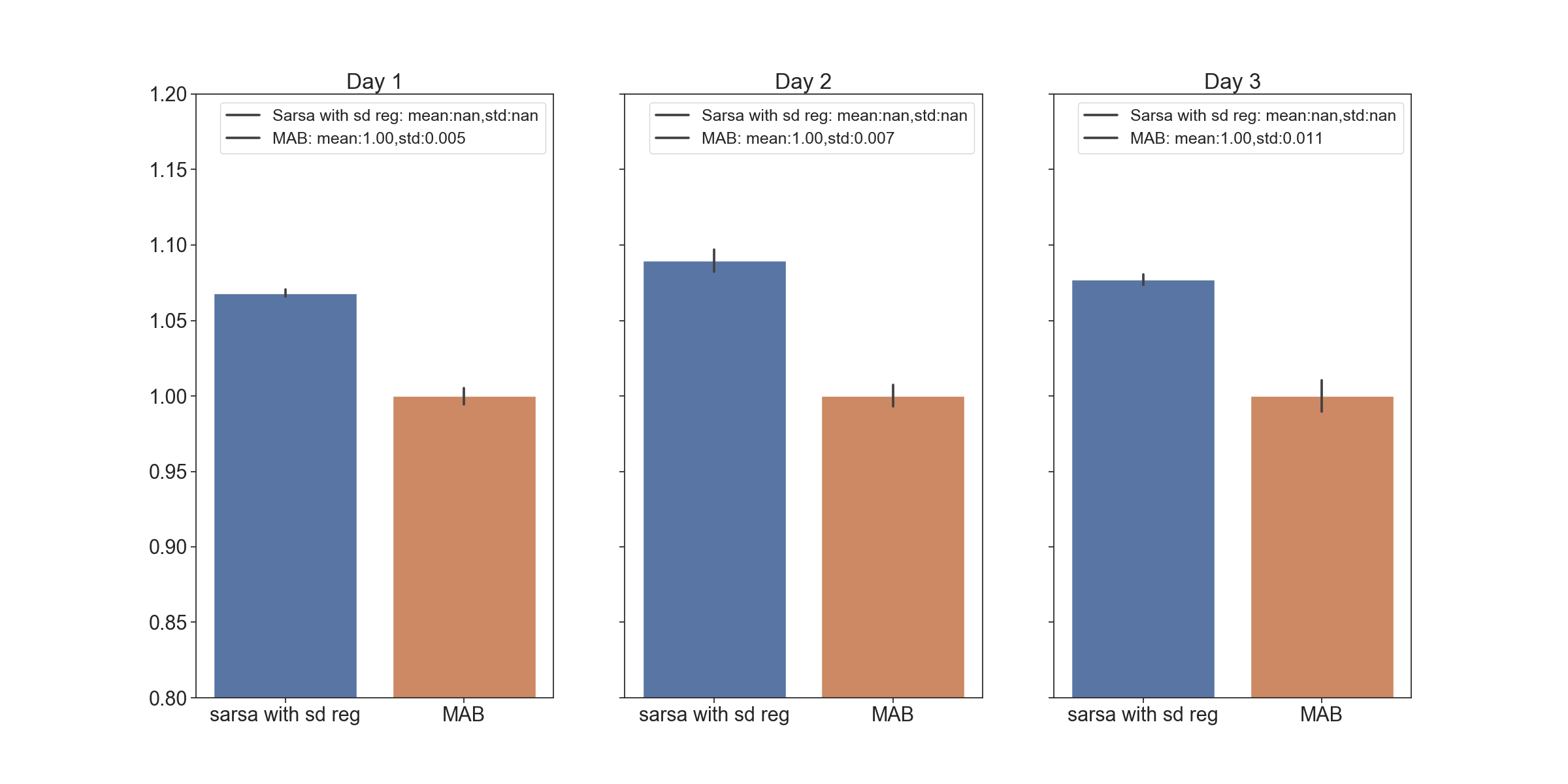}
	\caption{IPH comparison between SARSA with SD regularization and MAB. Results are standardized with respect to the average driver IPH for the MAB.}
	\label{fig:sd_reg_mab.png}
\end{center}
\end{figure}

\subsection{Benchmarking with Baseline}
\begin{figure}
\begin{center}
        \includegraphics[width=0.7\linewidth]{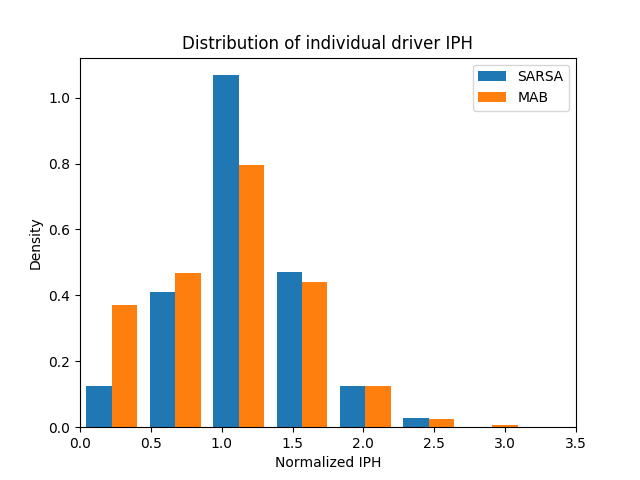}
	\caption{Individual driver IPH distribution comparison between SARSA with SD regularization and MAB. Results are standardized with respect to the average individual driver IPH using MAB.}
	\label{fig:driver_IPH.png}
\end{center}
\end{figure}

In this section, we compare SARSA (with all the algorithmic elements presented in Section \ref{sec:action-val}) to the MAB algorithm discussed in Section \ref{sec:benchmarking}. VPS (with or without stochastic policies) has a much higher run-time latency than these two methods as we explained in Section \ref{sec:action-val}. Since it does not meet the production latency requirement in the large-fleet scenario, we have excluded it from the comparison. Figure \ref{fig:sd_reg_mab.png} shows that the average IPH of SARSA is over 5\% higher than that of MAB across the different days. In Figure \ref{fig:driver_IPH.png}, we look at the IPH distributions of the two algorithms. The group with SARSA shows a more Gaussian distribution, whereas the distribution for the MAB group is skewed towards the lower side, partly explaining the difference in the means. 

\section{Real-world Deployment}
We have developed an AI Driver Assistant application powered by our proposed repositioning algorithms. We designed and ran a field experiment program on DiDi to evaluate the performance in the real-world environment. The goal is to compare VPS with human expert repositioning strategies, which is significantly more challenging than the simulation baselines. 
Human drivers have obtained the domain knowledge through experience and the real-time demand information provided by the app.

Unlike operating on an autonomous ride-hailing platform, testing a vehicle repositioning algorithm with human drivers in our case requires additional consideration of the way the repositioning recommendations are delivered and the drivers' willingness to accept and follow the recommendations, because the drivers within the experiment program were on a voluntary basis in terms of executing the repositioning tasks, due to various practical constraints.

Since our experiment's goal is to benchmark algorithms on long-term cumulative metrics (daily income rate) and the supply-demand context could vary significantly within a day, it would be ideal if all the drivers in the program are online for the same period time which is also sufficiently long, and the drivers always follow the repositioning recommendations, for a fair comparison.  We designed an incentive scheme to encourage the drivers to participate as closely to the ideal situation as possible.  Specifically, we required that they are online for at least five hours out of the eight hours from 11am to 7pm (the experiment interval) during weekdays and that they skip or fail to complete no more than three tasks each day.  The drivers were rewarded for each repositioning task that they finished, and they received additional reward for each day that they met the daily requirements. Income rates are computed using the data from the drivers who have met the daily requirements.  The form and level of the incentives certainly would influence the drivers' willingness to follow the repositioning tasks, and the associated elasticity apparently depends on the specific driver and task. We leave the investigation on the optimal incentive design in this context for future works.

\subsection{AI Driver Assistant}
\begin{figure}
\begin{center}
        \includegraphics[width=0.8\linewidth]{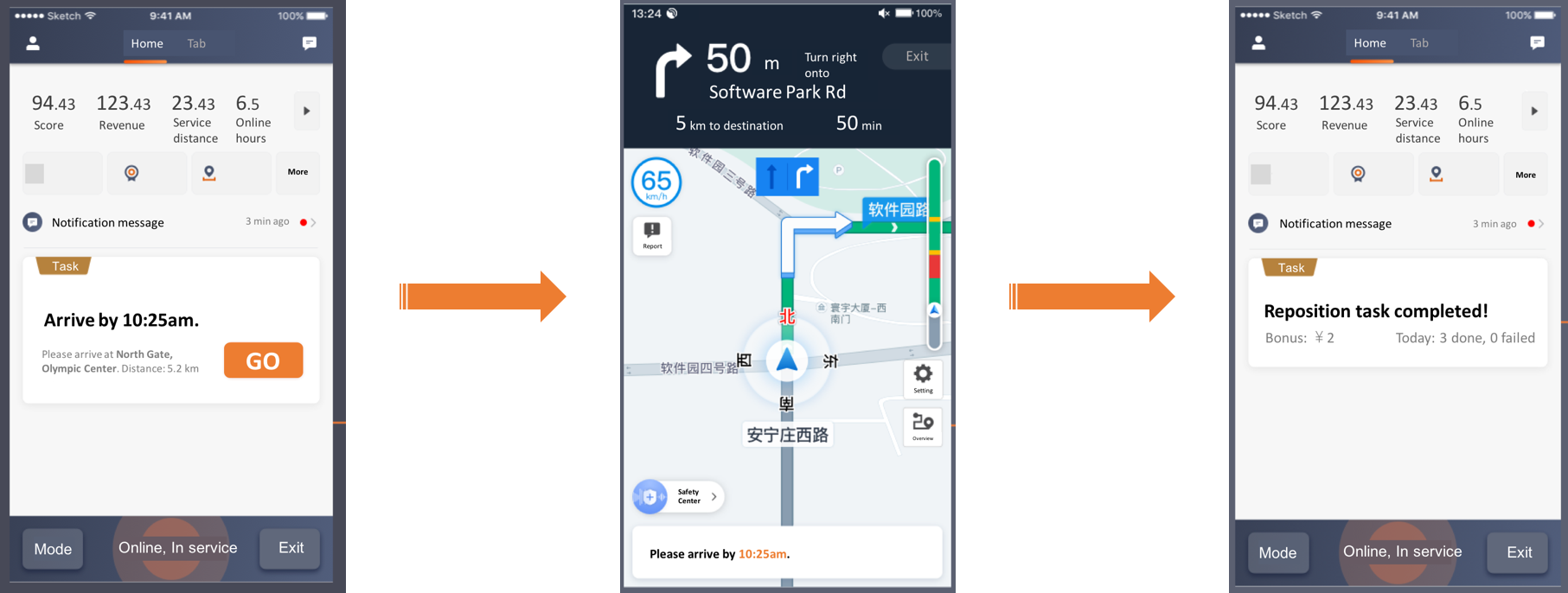}
	\caption{Illustrative screenshots of the AI Driver Assistant program. The screenshot on the left shows that the driver receives a repositioning task. The middle one shows the GPS navigation. When the driver completes the task, the app displays a confirmation (right).}
	\label{fig:scrrenshots}
\end{center}
\end{figure}

Repositioning recommendations are delivered through pop-up message cards within the mobile AI driver assistant app.  Once repositioning is triggered, a message card appears at the target driver's app.  The message card contains instructions for the repositioning task, including the destination and the target time that the driver is required to be there.  After the driver acknowledges the task, GPS navigation is launched to provide turn-by-turn route guidance to the driver.  The system automatically determines if the driver has reached the prescribed destination within the required time frame. Figure \ref{fig:scrrenshots} shows a few illustrative screen shots of the app. This app is integrated into the driver-side of the rideshare application and can be activated/deactivated on an individual basis.

\subsection{Small Fleet}\label{sec:real_small}
We recruited 61 experienced drivers across the same three cities as in the simulation experiments to participate in our experiment, which lasted for two weeks.  
The drivers were randomly divided into two groups of similar sizes.  The \emph{algo} group received repositioning recommendations from our proposed algorithm.  The 
\emph{human} group did not receive any recommendation, and they were free to determine their own cruising plan while idle.
Order dispatching was handled by the platform and was exogenous to both groups.  

\begin{figure}
\begin{center}
        \includegraphics[width=0.6\linewidth]{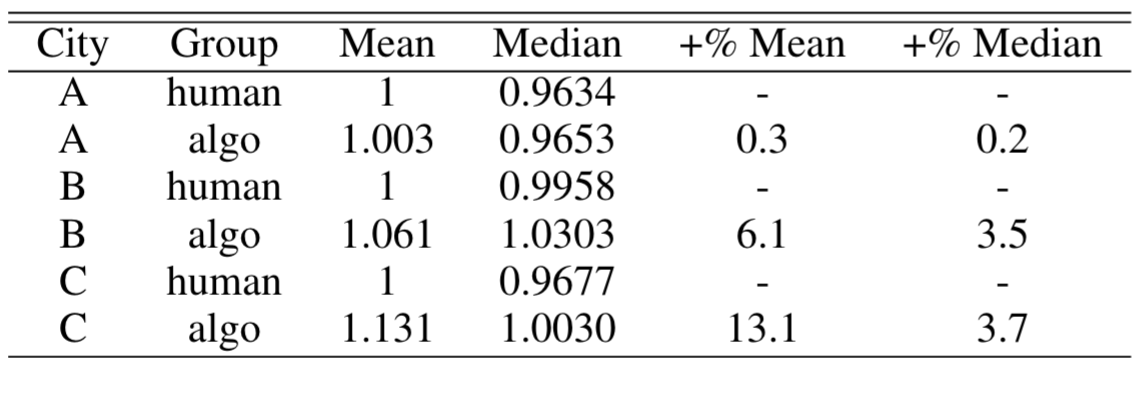}
	\caption{Daily income rate comparison for regular drivers without preferential order matching in the real-world experiment, normalized w.r.t. the means of the \emph{human} group.}
	\label{fig:three-city-iph}
\end{center}
\end{figure}

First, we look at the income rates across the three cities.  We have found that the advantage of algorithmic repositioning appears most significant for the group of regular drivers without preferential order matching.\footnote{Preferential order matching is typically applied to drivers with high service scores as incentives.}  Figure \ref{fig:three-city-iph} shows daily income rate comparison between the \emph{algo} group and the \emph{human} group by city.  Cities B and C both show significant improvement in income rate, with a difference of 6.1\% and 13.1\% in the mean respectively.  There is also a relative improvement of more than 3\% in the median for both cities. The improvement in city A is smaller than the other two cities because its demand is more abundant with respect to the supply (see Table \ref{tab:small_fleet_cities} in Appendix \ref{sec:app_smallfleet}), leading to less advantage for a repositioning algorithm - the drivers are less likely to idle and hence there is less room for improvement.
Over the entire group of drivers in the program, the relative improvement of the \emph{algo} group over the \emph{human} group is up to 2.6\% across the three cities.  We surmise that regular drivers without preferential order matching were more likely to be in need of idle-time cruising guidance as they have a higher chance of becoming idle during the day, thus offering larger room for improvement through algorithmic repositioning. 

Since our participation requirements to the drivers were voluntary and incentive-based, all the drivers might not have accepted or completed all the repositioning tasks that our algorithm issued.  We analyzed the \emph{algo} group data with more than five hours of daily online time and compared daily income rates of the  drivers who followed repositioning instructions closely to those of whom failed.  We have found that those drivers following the algorithmic repositioning faithfully enjoy an average income rate 9\% (4\% for median resp.) higher than those not, corroborating the effectiveness of our algorithm.

\subsection{Large Fleet}
To evaluate the performance of our algorithm on larger driver groups, we recruited about 1200 drivers (300 each for cities K and F, and 600 for city M) from three cities on the DiDi platform to participate in our pilot programs. We have shown in Section \ref{sec:sim-stoch} that coordination can no longer be ignored for a driver group of this size. Hence, we used the learning method for action values developed in Section \ref{sec:action-val}.

Different from the experiment for small driver groups, the drivers recruited in this case were to form the experiment group only. We selected the control pool directly from the general driver population. The control pool is constructed carefully so that the distributions of several key features, e.g., service scores, daily average online duration, are identical to those of the experiment group. The differences in IPH and utilization between the experiment group and control pool are also statistically insignificant. \footnote{For the experiment in city M, the control group was randomly selected from the drivers who showed interest to participate and were simply kept as an observation group during the experiment.}

The potential network effect between the experiment and control groups is mitigated in two ways. First, despite a much larger algorithmically managed fleet, the experiment group still accounts for a minor portion of the total vehicle population (about 1\% for cities K and F, and about 11\% for city M). Hence, the interaction of the two groups of vehicles is likely limited. Second, the evaluation method that we adopt (described in the next section) considers the performance of a large number of random group samples from the control pool, which are diverse in their degree of potential network effect with the experiment group depending on their spatiotemporal states. Our comparison is done against the distribution of performances over these samples.

\subsubsection{Evaluation Methodology}\label{sec:eval}
Unlike the experiment for small driver groups, where we compute individual driver performance and compare the distributions of the two groups, here the goal of the experiment is to compare group level performance. It should be noted that it is incorrect to determine the statistical significance of the results by computing the within-group performance variability because it is affected by the different individuals' characteristic profiles, whereas in comparing group-level performance, we care about the variability due to different samples of groups. In our case, we perform the significance test by bootstrapping. We treat the experiment (`algo') group as one sample $X$, which gives the group income rate $p(X)$ as in \eqref{eq:agg_iph}.
Then, we sample with replacement $N$ control groups of size $|X|$ from a control pool whose size is several times of $|X|$.  We denote each control group sample $\xsupk{Y}{n}, n=1,\cdots, N$. Then, we compute the 95\% confidence interval of the control group performance $\mathcal{C}_Y$ based on the set $\{p(\xsupk{Y}{n})\}_{n=1}^N$ and conclude that the performance difference is significant if $p(X)$ falls outside $\mathcal{C}_Y$. We tested different values of $N$ and have selected $N=5000$ when the confidence interval converges.

\subsubsection{Results}
We present the aggregated performance of the experiment group drivers by comparing it to that of a large number of sampled control groups from the general driver population as detailed in Section \ref{sec:eval}. We report results on group-level IPH \eqref{eq:agg_iph} as well as \emph{utilization}, which is defined as 
\begin{equation}
    u(X) := \frac{\sum_{x\in X}s(x)}{\sum_{x\in X}h(x)},
\end{equation}
where $s(x)$ is the total time in service for a driver $x$ over the course of the experiment. Utilization is the complement of idle time ratio. It is another important measure to quantify the efficiency of a ride-hailing platform. 

\begin{figure}
\begin{center}
    \includegraphics[width=0.6 \linewidth]{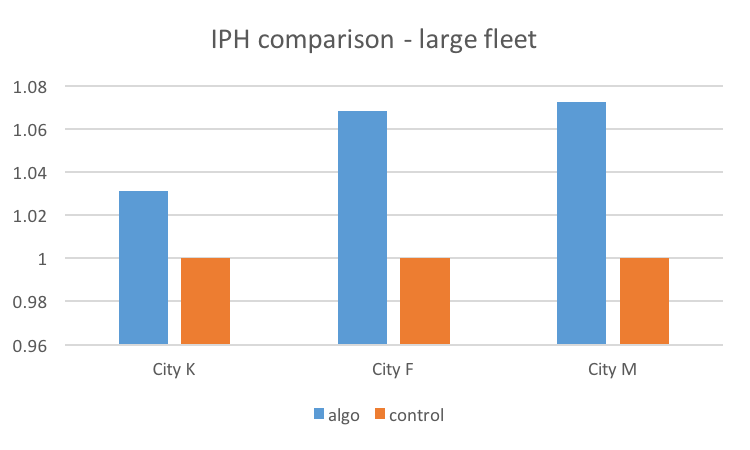}
    \includegraphics[width=0.6 \linewidth]{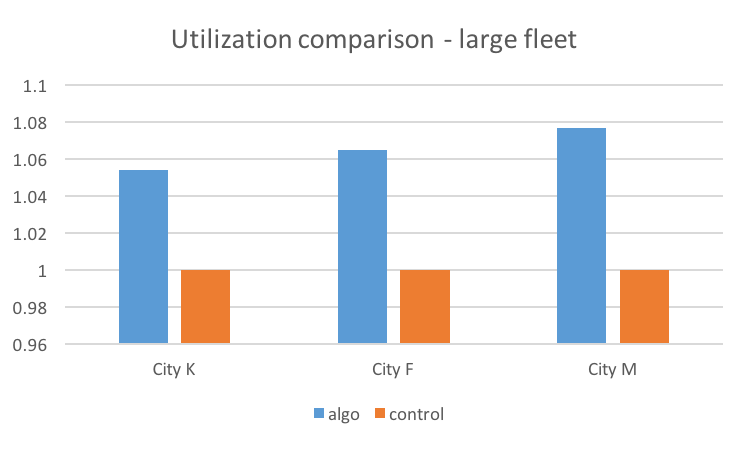}
	\caption{Income efficiency and utilization comparison for the large-fleet real-world experiment. All figures are standardized with respect to those of the control group. The 95\% confidence intervals for the control group are also plotted.}
	\label{fig:income_comp}
\end{center}
\end{figure}

The improvement in aggregated IPH and utilization for both cities is quite impressive as shown in Figure \ref{fig:income_comp}, with 3.1\% and 6.8\% increase in IPH for city K and F respectively, and 3.5 and 3.8 percent points increase in utilization respectively. These results are all statistically significant through the bootstrap sampling tests discussed in Section \ref{sec:eval}.

\begin{figure}
\begin{center}
    \includegraphics[width=0.45 \linewidth]{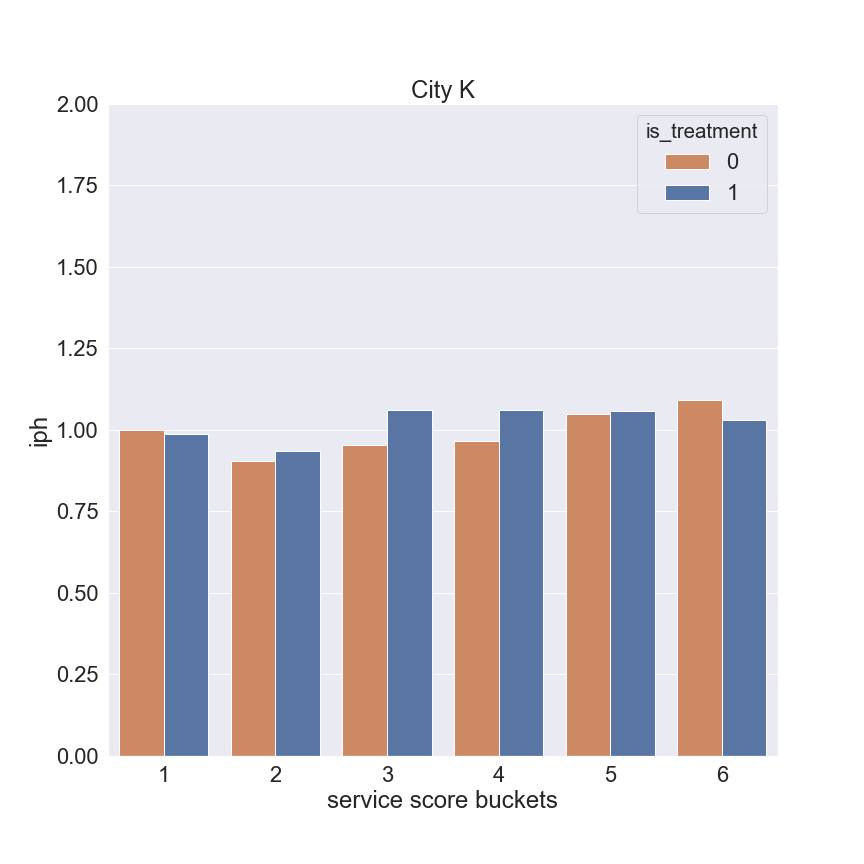}
    \includegraphics[width=0.45
    \linewidth]{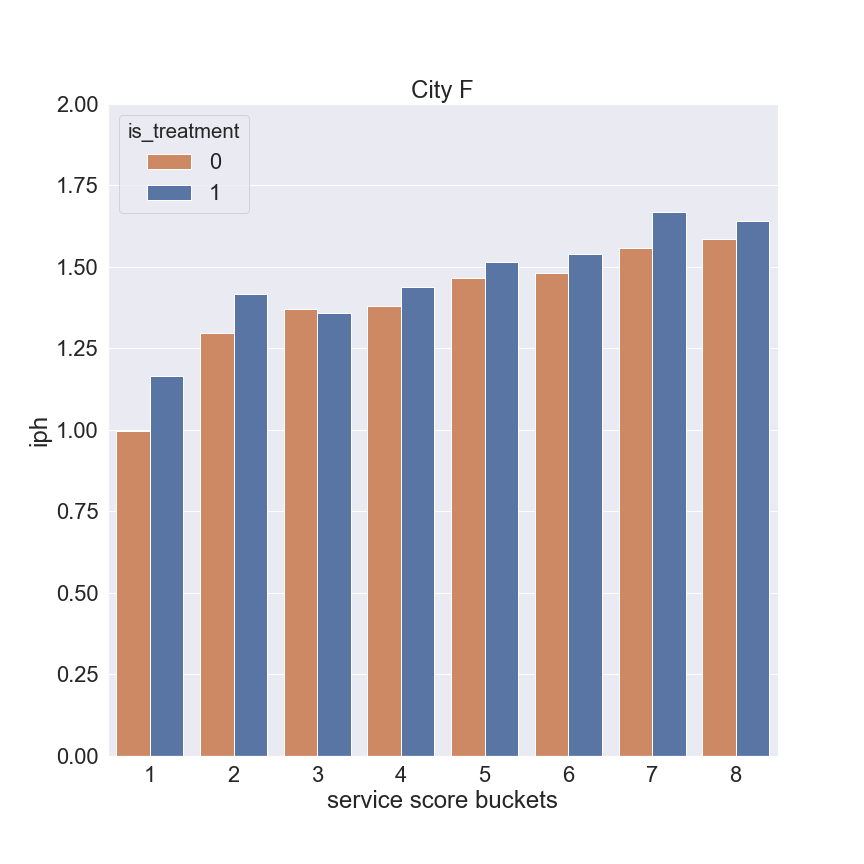}
    \includegraphics[width=0.45
    \linewidth]{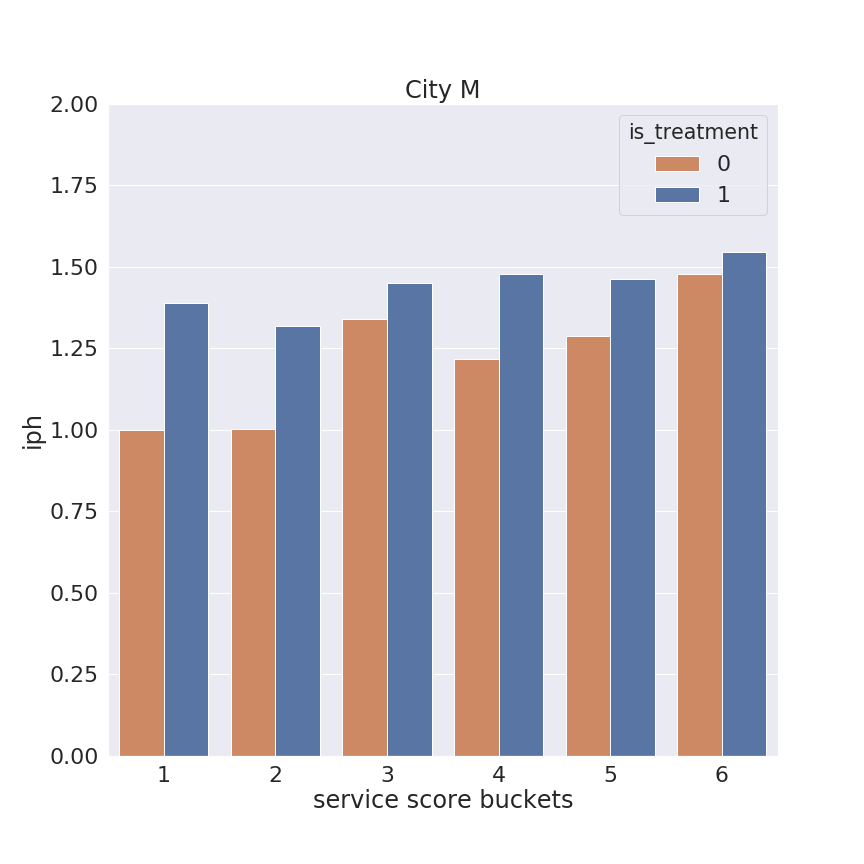}
	\caption{Income efficiency comparison by service score buckets. The value intervals of the buckets are mutually exclusive and cover the entire range of service scores in each city.}
	\label{fig:iph_over_kb}
\end{center}
\end{figure}

We further analyzed the performance comparison by dividing the drivers into \emph{service score} buckets. Drivers typically receive different levels of preferential matching according to their service scores, so it is an important factor contributing to the income variations within the experiment and control groups. Figure \ref{fig:iph_over_kb} shows the bar plots of IPH by service score buckets, which are arranged in increasing order of the score from the left to the right. We observe that except for two high score buckets, the experiment group consistently outperformed the control groups across all the buckets.

\begin{figure}
\begin{center}
    \includegraphics[width=0.45 \linewidth]{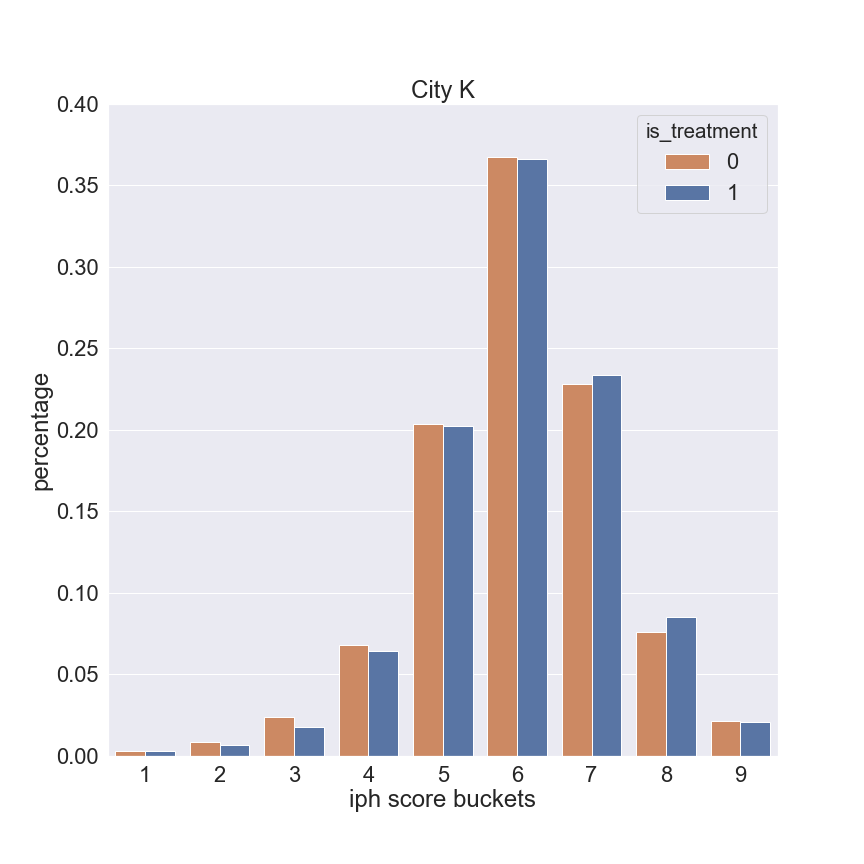}
    \includegraphics[width=0.45
    \linewidth]{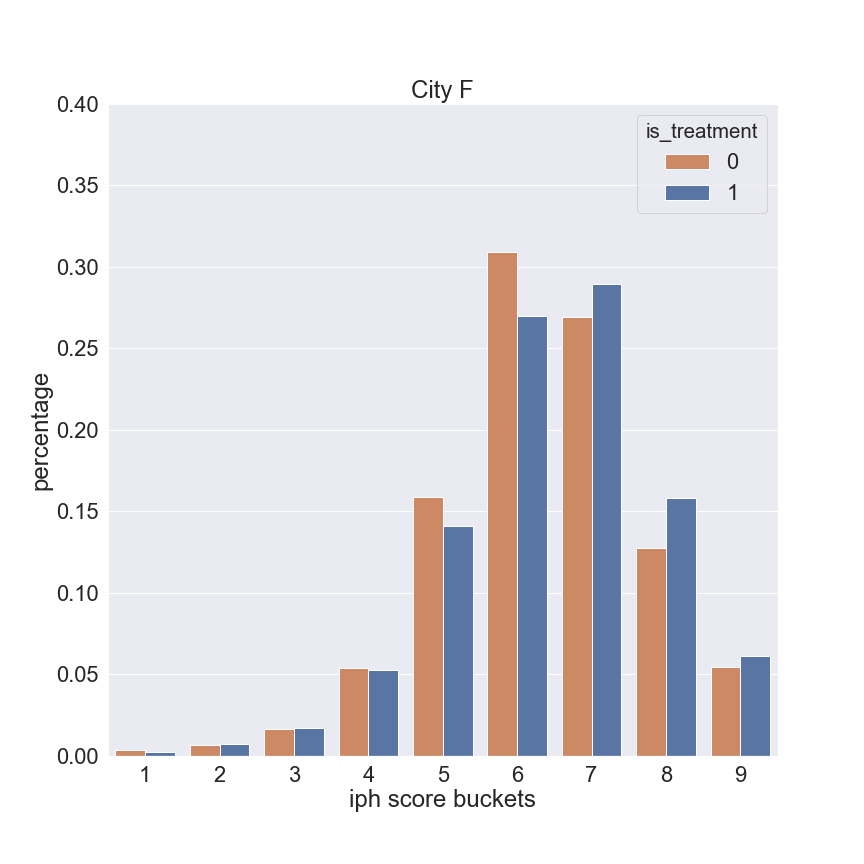}
    \includegraphics[width=0.45
    \linewidth]{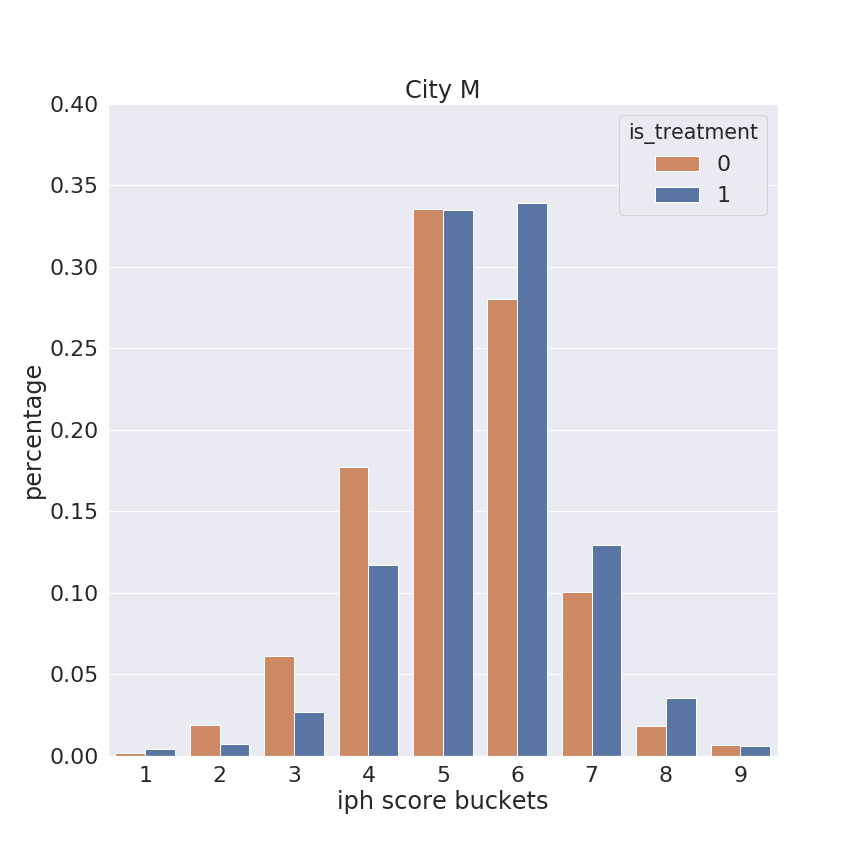}
	\caption{Individual-level IPH distribution comparison by IPH buckets. The value intervals of the buckets are mutually exclusive and cover the entire range of IPH in each city.}
	\label{fig:ind_iph_buckets}
\end{center}
\end{figure}

Although our primary metrics in the large-fleet experiments are at group level, we also compare individual-level IPH's in terms of their distributions for the two groups. Figure \ref{fig:ind_iph_buckets} shows the individual IPH distributions in the form of histograms over discretized and equally spaced IPH buckets, arranged in an increasing order. The results are apparently consistent with those of the group-level metrics. For each city, we see that there are generally more algo group drivers in the higher IPH buckets and fewer algo group drivers in the lower buckets than their counterparts from the control group. The difference in the IPH distribution is larger for cities F and M, which also echos with the group-level results in Figure \ref{fig:income_comp}. Another observation is that both groups have their IPH values distributed over multiple buckets, exhibiting in-group variations and thus empirically justifying the evaluation method in Section \ref{sec:eval}.

\subsection{Long Search}
From the experiment data, we have observed that more than 80\% of the time, the experiment group drivers do not need more than two consecutive repositioning actions before getting matched to an order.  This is echoed by further observation from analysis of the driver trajectories that long-search was rarely triggered during the experiment, implying that the drivers rarely went a long time without being dispatched to an order. This demonstrates that our general search algorithm could handle most of the cases successfully. 

\begin{figure}
\begin{subfigure}{.3\textwidth}
    \centering
	\includegraphics[width=0.8 \linewidth]{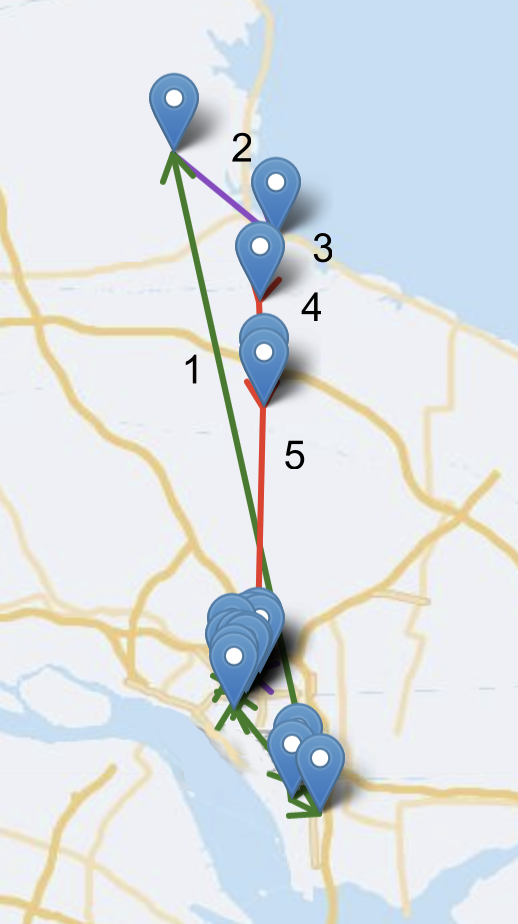}
	\caption{}
	\label{fig:LSa}
\end{subfigure}
\begin{subfigure}{.7\textwidth}
    \centering
    \includegraphics[width=0.9 \linewidth]{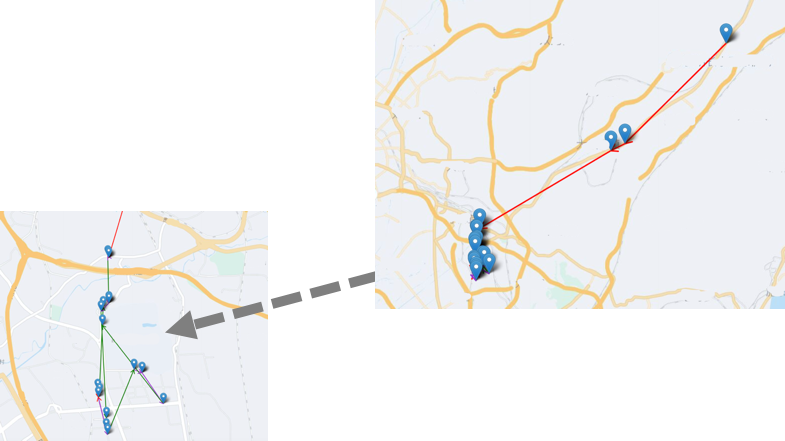}
    \caption{}
    \label{fig:LSb}
\end{subfigure}
\caption{Examples of using long-search to reposition a driver from large low demand area to a higher value region. Green arrow: order dispatching. Purple arrow: self-reposition. Red arrow: algorithmic reposition.}
\label{fig:LS}
\end{figure}

However, we do have cases when long-search has been particularly useful, as shown in Figure \ref{fig:LS}. The driver was brought by a long-distance order to the outside of the city in Figure \ref{fig:LSa}.
In this case, the driver self-repositioned once after finishing the long order without getting matched to another order. 

After two regular search steps (labels 3 and 4), long-search (label 5) was triggered and the search algorithm managed to find the right direction to reposition the driver right back to the city center.

In Figure \ref{fig:LSb}, the driver was at a rural area far from a town at the bottom left corner of the map. Two long-search repositions were triggered and guided the driver successfully to the town center, where consecutive orders were matched to the driver. (See inset at the left.) In both cases above, long-search prevented the drivers from being stuck in large low-value areas with targeted destinations of higher earning prospect, thus helping to improve their overall income efficiency. 

A participants survey was conducted after the experiment program and the feedback showed that the overwhelming majority of the drivers rated the program positively and would like to participate again.

\section{Discussion}
In this paper, we have considered two paradigms of vehicle repositioning, small and large fleets, which we have observed distinct enough to demand different algorithm designs. We focus on individual performance for small-fleet problems, where the vehicles under algorithmic guidance are essentially independent agents and their actions do not have significant impact on the supply-demand context. For large-fleet problems, we care about group level performance, and hence, coordination among the agents' actions becomes important. In this case, the SD information not only allows for a better characterization of the state, more responsive to the changes in environment, but also induces awareness of the neighboring agents in the policy at planning stage. 

Our approach does not explicitly depend on the type of MoD, e.g., single occupancy, carpooling. Our vehicle repositioning agent is equally applicable to an MoD platform with the presence of carpooling and other modes of ridesharing, as long as appropriate transition experience data is available. In fact, in our real-world deployment, the drivers are free to accept carpooling orders in addition to the regular ride-hailing trips. To specifically optimize for the carpool mode, more state features (e.g., number of passengers on board, the destination for each passenger) can be added to the Q-value network as done in \cite{al2019deeppool}.

We have assumed that the reposition tasks are always executed by the drivers. In practice, this happens only up to an acceptance probability $p_a$. There are a number of reasons for which the driver would reject the task, e.g., perceived travel distance, the need for a break. It is reasonable to think that $p_a$ would depend on both the driver's state and the reposition task, as well as some environment factors, e.g., weather, traffic. One possible simple way to account for $p_a$ in our algorithm is to multiply the estimate of the probabilities to the action-value vector $\bold{q}$ in \eqref{eq:softmax} so that the learned values are properly discounted by the corresponding chances of task rejection. Nevertheless, developing a prediction model that accurately learns $p_a$ remains a challenging task, and it requires more in-depth treatment for handling the resulting uncertainty.

Our current approach comprises an offline learning stage and an online planning stage. The value networks are fixed at run-time. For a highly dynamic environment like ride-hailing, it is natural to hope that online on-policy updates would make the value function more adaptive, thus leading to better performance. We leave the investigation in this direction to future works.

\subsubsection*{Acknowledgments}
The authors would like to thank Jing He, Jing Chen, Yuanfang Wang, Shaohui Su, Chunyang Liu, Zhe Xu, Ben Wang, Bin Liu, Chen Sang, Chengyi Shi, the DiDi regional operations teams, and other sister engineering teams for their support and efforts in product, platform, and operations for the experiment program of this research.

\bibliographystyle{abbrv}
\bibliography{tony_bib}

\appendix
\section{The DPE Algorithm}\label{sec:app_dpe}
For the purpose of implementation, we introduce a new binary option $b$ in place of $o$ under the same semi-MDP framework, with 0 indicating idle and 1 for dispatched.
$V(s | dispatch)$ and $V(s | idle)$ can then be represented as $V(s | b=1)$ and $V(s | b=0)$, respectively. The required conditional value function $V(s|o_d)$ is given by $V(s|b=1)$.
The main procedure of DPE is presented in Algorithm \ref{alg:dpe}.  
We use the same discount factor $\gamma = 0.92$ in DPE (for computing the state values) and in VPS (for computing the path values).  
For the cerebellar embedding we use 3 quantization functions and a memory size $A$ of 20000. The embedding dimension $m$ is chosen to be 50.
Following the cerebellar embedding layer are fully connected layers having [32, 128, 32] hidden units with ReLU activation.
To evaluate the policy we apply Adam optimizer with a constant step size $3e^{-4}$ and the Lipschitz regularization parameter $\lambda$ as $1e^{-4}$.

\begin{algorithm}
\caption{Dual Policy Evaluation (DPE) with Cerebellar Value Network (CVNet)}
\begin{algorithmic}[1]\label{alg:dpe}
\STATE Given:
historical driver trajectories $\{(s_{i,0}, o_{i,0}, r_{i,1}, s_{i,1}, o_{i,1}, r_{i,2}, ..., r_{i,T_i}, s_{i,T_i})\}_{i \in \mathcal{H}}$ collected by executing a (unknown) policy $\pi$ in the environment.
\STATE Given: $n$ cerebellar quantization functions $\{q_1, ..., q_n\}$, regularization parameter, max iterations, embedding memory size, embedding dimension, memory mapping function, discount factor $\lambda, N, A, m, g(\cdot), \gamma$.
\STATE Compute training data from the driver trajectories as a set of (state, reward, next state) tuples, e.g., $\{(s_{i,t}, R_{i,t}, s_{i,t+k_{i,t}})\}_{i \in \mathcal{H}, t=0,...,T_i}$ where $k_{i, t}$ is the duration of the trip.
\STATE Initialize the embedding weights $\theta^M$.
\STATE Initialize the state value network $V(s)$ with random weights $\theta_1$.
\STATE Initialize the conditional state value network $V(s|b)$ with weights $\theta_2$.
\FOR{$\kappa = 1, 2, \cdots, N$}
  \STATE Sample a random mini-batch $\{(s_{i,t}, R_{i,t}, s_{i,t+k_{i,t}})\}$ from the training data.
  \STATE Embed the states $s_{i, t}$ and $s_{i, t+k_{i,t}}$ using the quantization functions $\{q_1, ..., q_n\}$ and the memory mapping function $g(\cdot)$, e.g., $s \leftarrow c(s)^T \theta^M / n$ where the activation vector $c(s)$ is initialized to 0 and iteratively adding 1 to the $g(q_i(s))$-th entry of $c(s)$ for $i = 1, ..., n$.
  \STATE Set the binary option indicating idle movement or dispatch $b \leftarrow \mathbf{1}_{R_{i,t} \geq 0}$
  \STATE Update the state value network $V(s)$ with inputs $s_{i, t}$ and targets $\frac{R_{i,t}(\gamma^{k_{i,t}} - 1)}{k_{i,t}(\gamma - 1)} + \gamma^{k_{i,t}} V(s_{i, t+k_{i,t}}) $.
  \STATE Update the conditional state value network $V(s|b)$ with inputs $[s_{i, t};b]$ and targets $\frac{R_{i,t}(\gamma^{k_{i,t}} - 1)}{k_{i,t}(\gamma - 1)} + \gamma^{k_{i,t}} V(s_{i, t+k_{i,t}}) $.
\ENDFOR
\RETURN $V$
\end{algorithmic}
\end{algorithm}

\section{Dispatch Probability Model}\label{sec:app_dispatchprob}
We describe the experiment configuration for the results in Table \ref{table:dispatch_prob}. For each city, we train a LightGBM
decision tree, and use it to predict the probability of a driver receiving trip request given his or her current state. State features include driver\_ids, driver's location, time, and day of the week. We hash and encode each driver\_id into a 5 dimensional dense vector, and treat each element of the vector as an input feature. Hyper-parameters are tuned via Bayesian optimization,
 where we fit a Gaussian process on target hyper-parameters, and use it as a surrogate model to optimize test AUC. We report the best hyper-parameter configuration in Table \ref{table:dispatch_params}.

\begin{table}[htp]\center
\begin{tabular}{cccccc}
\hline
City & Recall & Precision & F1 & Accuracy & AUC \\\hline
A &0.7782 &0.7596 &0.7835 &0.8014 &0.876\\
B &0.7568 &0.7592 &0.7618 &0.7761 &0.853\\
C &0.7745 &0.7834 &0.7812 &0.7977 &0.8729\\\hline
\end{tabular}
\caption{Evaluation results of the dispatch probability models.}
\label{table:dispatch_prob}
\end{table}

\begin{table}[htp]\center
\begin{tabular}{cccccc}
\hline
boosting\_type & num\_leaves & max\_bin  \\\hline
gbdt & 6500& 8500\\\hline
min\_data\_in\_leaf & lambda\_l2 & learning\_rate\\\hline
26 &0.01846772 &0.0834044 & \\
\end{tabular}
\caption{Hyper-parameter configuration of the dispatch probability models.}
\label{table:dispatch_params}
\end{table}

\section{Simulation Environment}\label{sec:app_sim}
We describe in detail the dynamics of the ride-hailing simulation environment and the associated data sets. 

\subsection{Dynamics}

The data generation of the simulator is based on replaying the historical trip request data of the particular dates. For each given date, the simulation environment is initialized by the drivers' states and the orders information per historical data at the start time of the simulation. The set of vehicles managed by the reposition policy are sampled from the initial group of drivers. 
The subsequent states of the vehicles are completely determined by the matching and repositioning operations, as well as the idle cruising and online-offline behavior models, which are trained on real data prior to the simulation date. The reposition-managed vehicles are always online and are independent from the behavior models. Specifically, the simulator batches idle drivers and open requests over matching windows of a constant length (typically a few seconds) and performs order-driver matching at the end of each window using the KM (Hungarian) algorithm on a bipartite graph of the drivers and passengers. The edge weights of the graph are determined by the order dispatching policy. With the minimum-distance policy, for example, the edge weight is the travel distance between the passenger and the driver at the request time, i.e., the pick-up distance. All the matched vehicles will then go to the request location and transport the passenger to the destination with the timestamp determined by the corresponding predicted travel time. With a probability determined by an order cancellation model, an order is cancelled based on the pick-up distance. Upon order completion, regular drivers move according to the idle cruising model to a neighboring hexagon grid cell, if there is no immediate new request matched to the drivers. For reposition-managed vehicles, they are batched in a similar way to matching (but with a longer interval, e.g., 100s) and the reposition destinations are determined by the reposition policy. Prior to the next round of matching, current drivers go offline and potential new drivers appear online according to the online-offline behavior model. 

We have released an open simulation platform\footnote{https://outreach.didichuxing.com/Simulation/} powered by environments with the same dynamics described above. The simulation engine has also supported the KDD Cup RL Track competition in 2020.

\subsection{Data}

The anonymized data sets used to power the simulator consists of four parts: vehicle trajectories, trip requests, idle cruising, and order cancellation. The vehicle trajectory data contains instances of timestamps and the corresponding vehicle location in coordinates for each combination of driver ID and order ID. The trip request data specifies for each order the start and end times, the pick-up and drop-off locations, and the reward units (in the sense of trip fee). The idle cruising model is trained on real data and specifies the transition probability for each pair of origin and destination grid cell for each hour of the day. Each OD pair is a potential idle cruising trip. The order cancellation model outputs the cancellation probability for a given value of pick-up distance. It captures the nonlinear cancellation behavior of a passenger in response to the pick-up distance once an order is matched. A sample of these data have been released through the DiDi GAIA Initiative.\footnote{https://outreach.
didichuxing.com/research/opendata/en/}

\section{Real-world Experiment: Small Fleet}\label{sec:app_smallfleet}
Table \ref{tab:small_fleet_cities} shows the basic background information of the three cities in the small-fleet experiments in Section \ref{sec:real_small}. The first column shows the relative amount of supply (available vehicles) with respect to the demand (passenger requests). The comparison is over the spatiotemporal dimensions instead of a simple comparison of the total numbers. For example, `scarce' means that during the experiment period, there are more zones (defined by hex cells) and time intervals where it is under-supplied than those over-supplied.

\begin{table}\center
\begin{tabular}{ccc}
\hline
City & Supply vs Demand & Population Size (millions)\\\hline
A & scarce & 4.51 \\
B & abundant & 7.31 \\
C & abundant & 4.65 \\\hline
\end{tabular}
\caption{The setup of the three pilot cities for the small-fleet experiments.}
\label{tab:small_fleet_cities}
\end{table}

\end{document}